\documentclass[11pt]{article}

\PassOptionsToPackage{table}{xcolor}
\usepackage[final]{acl}

\usepackage{times}
\usepackage{latexsym}

\usepackage[T1]{fontenc}

\usepackage[utf8]{inputenc}

\usepackage{microtype}

\usepackage{inconsolata}

\usepackage{graphicx}

\usepackage{amsmath}
\usepackage{amsthm}
\usepackage{amsfonts}
\usepackage{amssymb}
\usepackage{booktabs}
\usepackage{algorithm}
\usepackage{algorithmic}
\usepackage{soul}
\usepackage{listings}
\usepackage[most]{tcolorbox}
\usepackage{cuted}
\usepackage{tabularx}
\usepackage{array}
\usepackage{pifont}

\definecolor{oursbg}{RGB}{235,245,255}
\newcommand{\cmark}{\ding{51}}
\newcommand{\xmark}{\ding{55}}

\lstdefinestyle{promptstyle}{
    basicstyle=\ttfamily\small,
    breaklines=true,
    breakatwhitespace=false,
    columns=fullflexible,
    keepspaces=true,
    showstringspaces=false,
    breakindent=0pt,
    breakautoindent=false,
    escapeinside={(*}{*)}
}

\title{World Models for Policy Refinement in StarCraft II}

\author{
\textbf{Yixin Zhang\textsuperscript{1,2,*}},
\textbf{Ziyi Wang\textsuperscript{1,2,*}},
\textbf{Yiming Rong\textsuperscript{1,2,*}},
\textbf{Haoxi Wang\textsuperscript{1,2}},
\textbf{Jinling Jiang\textsuperscript{1,2}},
\\
\textbf{Shuang Xu\textsuperscript{1}},
\textbf{Haoran Wu\textsuperscript{1,\dag}},
\textbf{Shiyu Zhou\textsuperscript{1,\dag}},
\textbf{Bo Xu\textsuperscript{1,2,\dag}}
\\[1mm]
\textsuperscript{1}The Key Laboratory of Cognition and Decision Intelligence for Complex Systems,\\
Institute of Automation, Chinese Academy of Sciences
\\
\textsuperscript{2}School of Artificial Intelligence, University of Chinese Academy of Sciences
\\
\texttt{\{zhangyixin2024, wuhaoran2018, shiyu.zhou, xubo\}@ia.ac.cn}
}

\begin{document}
\maketitle

\begingroup
\renewcommand{\thefootnote}{\fnsymbol{footnote}}
\footnotetext[1]{These authors contributed equally.}
\footnotetext[2]{Corresponding authors.}
\endgroup
\begin{abstract}
Large Language Models (LLMs) have recently shown strong reasoning capabilities, motivating their use in complex decision-making environments. StarCraft II (SC2), with its massive state-action space and partial observability, is a challenging testbed. However, existing LLM-based SC2 agents primarily focus on improving the policy itself, leaving the integration of a learnable, action-conditioned dynamics model into the decision loop largely unexplored. In this work, we propose \emph{StarWM} and \emph{StarWM-Agent}, and conduct the first systematic study of the learnability and decision utility of player-view, action-conditioned textual world models for SC2. StarWM predicts short-horizon future observations under partial observability. StarWM-Agent integrates StarWM into a lightweight \emph{Generate--Simulate--Refine} loop for foresight-driven policy refinement. Extensive experiments show that StarWM substantially outperforms zero-shot baselines across multiple dimensions, while StarWM-Agent achieves consistent win-rate gains of 30\%, 15\%, and 30\% against the SC2 built-in AI at Hard (LV5), Harder (LV6), and VeryHard (LV7), respectively. Additional analyses show that our method is complementary to existing history-summarization SC2-agent approaches, with their combination achieving the strongest online performance in our evaluation. Code is available at \url{https://github.com/yxzzhang/StarWM}.
\end{abstract}

\section{Introduction}
In recent years, Large Language Models (LLMs) have demonstrated remarkable reasoning and generalization capabilities, extending their utility from general language tasks to complex decision-making scenarios~\cite{yao2023react,schick2023toolformer,wang2023voyager}. StarCraft II (SC2), with its enormous state-action space and imperfect information setting, serves as an ideal environment for testing the complex decision-making capabilities of LLMs. Recent work has explored LLM-based SC2 agents from multiple angles, including observation summarization mechanisms~\cite{textstarcraft2}, augmenting inputs with external knowledge and multi-modal features~\cite{LLMPySC2SI}, and hierarchical designs that separate strategic planning from tactical execution~\cite{SC2Arena}. Despite these advances, most methods primarily focus on improving the \emph{policy} itself, while leaving the integration of a learnable, action-conditioned dynamics model into the decision loop largely unexplored.

Cognitive science research indicates that when addressing complex tasks, humans often rely on internal world models to perform short-term simulation, adjusting their actions to avoid penalties and maximize rewards~\cite{buildmachineslikepeople}. In machine learning, world models have been extensively studied across multiple domains, such as model-based RL~\cite{mbpo,muzero,dreamerv3} and autonomous driving~\cite{drivewm,GAIA2}. However, to the best of our knowledge, there is still no systematic study of world models for SC2. Earlier attempts such as StarCraft Defogger~\cite{starcraftdefogger} focus on \emph{state extrapolation} ($P(s_{t+k}\mid o_{\le t})$) rather than \emph{action-conditioned dynamics modeling} ($P(s_{t+k}\mid o_{\le t}, a_t)$), and thus cannot serve as a forward simulator for policy refinement.

\begin{figure*}[tb]
    \centering
    \includegraphics[width=\linewidth]{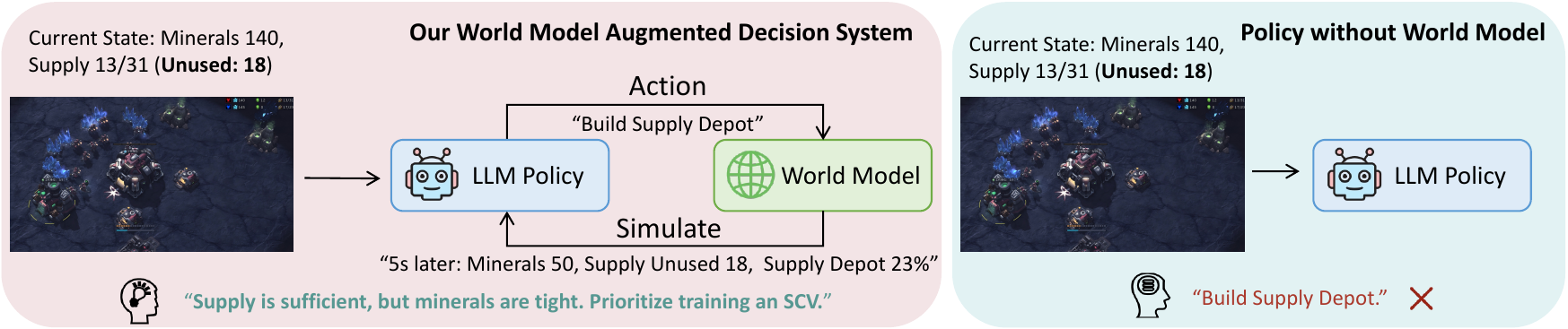} 
    \caption{Case study comparing our world-model-augmented decision system (StarWM-Agent) with a policy that does not use a world model. Given the current observation, the LLM policy initially proposes \emph{Build Supply Depot}. A 5-second rollout by the world model predicts that minerals will drop to 50 and the supply depot will be 23\% complete, while unused supply remains 18. Based on this prediction, the system revises the action to \emph{Train SCV}, avoiding premature infrastructure expenditure that would lead to mineral shortage. This example illustrates that incorporating a world model can improve macro-management decision-making.}
    \label{fig:online_casestudy}
\end{figure*}

We attribute the lack of SC2 world models to two major challenges. First, \textbf{learning SC2 dynamics is inherently challenging}: the environment exhibits strongly coupled hybrid dynamics, involving resource flows, task progression, micro-level unit kinematics, and combat evolution governed by damage mechanics, all under partial observability. Second, \textbf{the decision utility of a learned world model is non-trivial}: even with a learned world model, it remains unclear how to turn predicted futures into better LLM decisions under a practical \emph{performance--cost trade-off}, without resorting to expensive search algorithms.

To address these challenges, we conduct the first systematic study of the \textbf{learnability and decision utility} of player-view, action-conditioned textual world models for SC2. We propose \textbf{StarWM}, a learnable action-conditioned dynamics model that predicts short-horizon future observations conditioned on the current observation and proposed actions under partial observability. Specifically, we introduce a textual representation that factorizes SC2 observations into five semantic modules. Based on this representation, we construct \textbf{SC2-Dynamics-50K}, the first instruction-tuning dataset for SC2 dynamics prediction, and train StarWM via supervised fine-tuning. To comprehensively evaluate the learned world model, we develop a multi-dimensional offline evaluation framework tailored to SC2's hybrid dynamics, covering economy, development, micro-entities, and macro-situation. To assess its decision utility, we develop \textbf{StarWM-Agent}, a world-model-augmented decision system that integrates StarWM into a lightweight \emph{Generate--Simulate--Refine} loop for foresight-driven policy refinement. Offline evaluation results show that StarWM achieves substantial improvements over zero-shot baselines, including nearly 60\% gains in resource prediction accuracy and self-side macro-situation consistency. Online tests of StarWM-Agent demonstrate consistent win-rate gains of 30\%, 15\%, and 30\% against the SC2 built-in AI at Hard (LV5), Harder (LV6), and VeryHard (LV7), respectively. Additional analyses show that action-conditioned forward simulation complements history summarization used in existing SC2 agents, with their combination achieving the strongest online performance in our evaluation.

In summary, our main contributions are threefold:
\begin{itemize}
    \item \textbf{Systematic Study of SC2 World Models:} We conduct the first systematic study of the learnability and decision utility of player-view, action-conditioned textual world models for SC2, providing empirical insights that delineate their capabilities and current boundaries.
    \item \textbf{Dataset and Evaluation Framework:} We construct SC2-Dynamics-50K, the first instruction-tuning dataset for SC2 dynamics prediction, and develop a multi-dimensional offline evaluation framework tailored to SC2's hybrid dynamics.
    \item \textbf{World-Model-Augmented Decision System:} We develop StarWM-Agent, which integrates StarWM into a lightweight \emph{Generate--Simulate--Refine} decision loop for inference-time policy refinement and yields consistent improvements in online decision-making.
\end{itemize}

\section{Related Work}

\subsection{World Models in Decision Making}
Research on world models in decision making focuses on modeling environment dynamics to support policy improvement. In model-based RL, works such as DreamerV3~\citep{dreamerv3} and MuZero~\citep{muzero} learn latent dynamics models for imagination-based optimization or online Monte Carlo Tree Search (MCTS), respectively. In autonomous driving, DriveWM~\citep{drivewm} and GAIA-2~\citep{GAIA2} use world models to generate high-fidelity video streams for online trajectory planning or offline long-tail data synthesis. Recent work has also explored using LLMs as world models in textual decision-making environments. RAP~\citep{hao2023reasoning} and WebDreamer~\citep{webdreamer} use LLMs to simulate action consequences, but typically rely on expensive planning algorithms such as MCTS or Model Predictive Control (MPC) for decision making. However, to the best of our knowledge, prior work has not studied the learnability and decision utility of world models for SC2, where partial observability, hybrid dynamics, and the huge state-action space make both dynamics modeling and cost-effective decision integration challenging.

\subsection{LLMs in StarCraft II}
The success of AlphaStar~\cite{alphastar} demonstrated the potential of end-to-end neural networks in SC2, but it relies on massive human data and league-based self-play training, with high computational costs and unverified out-of-distribution generalization. Recent work explores leveraging the pre-training knowledge of LLMs to build more generalized and interpretable agents with lower training cost. Existing studies explore enhancing LLM decision performance from various dimensions: TextStarCraft II~\cite{textstarcraft2} proposes Chain of Summarization to compress observation histories; LLM-PySC2~\cite{LLMPySC2SI} introduces external knowledge and multi-modal observations. StarEvolve~\cite{SC2Arena} employs a hierarchical framework to decouple strategic planning and tactical execution. Comm-CoT~\cite{CommCoT} achieves task decomposition via multi-agent collaboration. MASMP~\cite{MASMP} introduces natural language state machines and strategic memory to constrain action generation.
However, most existing methods focus on enhancing the LLM-based policy itself and have not systematically explored introducing a learnable dynamics model into the decision loop.

\section{Method}

\subsection{Problem Modeling}
We model SC2 as a Partially Observable Markov Decision Process (POMDP), represented as a tuple $\langle \mathcal{S}, \mathcal{A}, \mathcal{T}, \mathcal{R}, \Omega, \mathcal{O}, \gamma \rangle$, where $\mathcal{S}, \mathcal{A}, \Omega$ denote the state, action, and observation spaces; $\mathcal{T}(s' \mid s,a)$, $\mathcal{O}(o \mid s',a)$ and $\mathcal{R}(s,a)$ denote the transition, observation, and reward functions; and $\gamma \in [0,1]$ is the discount factor. Under this formulation, it is crucial to distinguish the environment simulator from the world model studied in this work.

\subsubsection*{Environment Simulator vs. World Model}
The \textbf{Environment Simulator (God View)}, such as the SC2 engine, operates on perfect information with access to the global state $s_t \in \mathcal{S}$. Its transition function $\mathcal{T}: \mathcal{S} \times \mathcal{A}_{\text{all}} \rightarrow \mathcal{S}$ calculates the next global state based on joint actions of all players. While it executes the environment's objective transition rules, it is inherently \emph{inaccessible} to any player constrained by the Fog of War.

\noindent The \textbf{World Model (Player View)} studied in this paper is built on a single player's restricted perspective, under the Fog of War and imperfect information. It can only access local observations $o_t \in \Omega$ and faces epistemic uncertainty regarding global states and opponent intentions. We define the world model as a probabilistic, action-conditioned dynamics model that learns the distribution of future observations, rather than full-state transitions.

\subsubsection*{Formulation of Task}
We formulate the world model's prediction task as follows: given the player's current observation $o_t$ and a sequence of intended actions $a_{t:t+\tau}$, the world model $\mathcal{M}_{\phi}$ aims to predict the future observation $\hat{o}_{t+\tau}$ after $\tau$ seconds. Formally:
\begin{equation}\label{equation:world model task}
\hat{o}_{t+\tau} \sim P_{\mathcal{M}_\phi}\!\left( o_{t+\tau} \mid o_t, a_{t:t+\tau} \right).
\end{equation}
This prediction task involves two key challenges:
\begin{itemize}
    \item \textbf{Self-side Evolution:} Predicting the opponent-independent parts of the future observation, such as self-side resources, supply, technology, and production progress.
    \item \textbf{Opponent-Dependent Interaction:} Implicitly inferring possible opponent actions and the opponent-dependent parts of the future observation, such as visible hostile units and combat outcomes.
\end{itemize}

\subsection{StarWM: A World Model for SC2}
\label{method:world model}

\paragraph{Textual Representation.}
The SC2 engine provides highly heterogeneous, structured information, including scalar values (e.g., minerals, gas, supply), categorical variables (e.g., unit types, upgrades), and spatial attributes (e.g., positions). To map them into an LLM-compatible text space, we serialize the engine-side observation $x_t$ into a textual observation $o_t$, and action $u_t$ into a textual action $a_t$,
\begin{equation}
o_t=\Phi(x_t), \quad a_{t}=\Psi(u_{t}),
\end{equation}
where $\Phi$ denotes the observation-to-text operator and $\Psi$ denotes the action-to-text operator.

Our key insight is that SC2 dynamics are inherently hybrid, governed by multiple distinct evolution mechanisms: resource changes are governed by accumulation and consumption; task progress follows deterministic temporal progression; unit movement follows spatial kinematics; and combat outcomes are driven by damage mechanics. Thus, we propose a textual observation representation, explicitly factorizing observation into five distinct semantic modules:
\begin{equation}
\begin{aligned}
\Phi(x_t)=\big[ &
f_{\text{info}}(x_t);\,
f_{\text{queue}}(x_t);\,
f_{\text{units}}(x_t); \\
&
f_{\text{struct}}(x_t);\,
f_{\text{hostile}}(x_t)
\big],
\end{aligned}
\end{equation}
where $f_{\text{info}}$ describes economy and status, $f_{\text{queue}}$ records ongoing tasks with progress, $f_{\text{units}}$ and $f_{\text{struct}}$ describe self units and structures with their attributes, and $f_{\text{hostile}}$ describes visible enemy entities. More details and examples of textual representation are provided in Appendix~\ref{appendix:Textual Representation}.

\paragraph{Training Dataset.}
Based on this representation, we construct \emph{SC2-Dynamics-50K}, a trajectory-based dataset for dynamics prediction. Specifically, we collect game trajectories using a rule-based scripted agent against the SC2 built-in AI, and convert each trajectory into textual training triples of the form $(o_t, a_{t:t+\tau}, o_{t+\tau})$, where $o_t=\Phi(x_t)$ and $o_{t+\tau}=\Phi(x_{t+\tau})$ are the current and future observations, and $a_{t:t+\tau}=\Psi(u_{t:t+\tau})$ is the action sequence executed from time $t$ to $t+\tau$.

\paragraph{Training Objective.}
We train the \emph{StarWM} by supervised fine-tuning (SFT) on SC2-Dynamics-50K. The training objective is the standard negative log-likelihood:
\begin{equation}
\mathcal{L}_{\text{SFT}}
=
-\sum_{\mathcal{D}}
\log P_{\mathcal{M}_\phi}\!\left(o_{t+\tau}\mid o_t,a_{t:t+\tau}\right),
\end{equation}
where $\mathcal{D}$ denotes the SC2-Dynamics-50K dataset. In this way, StarWM learns to model action-conditioned dynamics and predict future observations in Eq.~\ref{equation:world model task} under partial observability.

\subsection{Multi-Dimensional Offline Evaluation Framework}\label{offlineevaluationsystem}
Existing text-overlap metrics such as BLEU~\cite{bleu} and ROUGE~\cite{rouge} fail to capture numerical accuracy, spatial consistency, and transition-level correctness. Thus, we propose a multi-dimensional evaluation framework tailored to SC2's hybrid dynamics. Given a pair of ground-truth and predicted observations $(o_t, \hat{o}_t)$, we parse them into structured variables and evaluate them along four aspects:

\paragraph{Economy \& Status.}
This dimension focuses on economy and status. We adopt the \emph{Symmetric Mean Absolute Percentage Error (SMAPE)} for numerical prediction:
\begin{equation}
\text{SMAPE} = \frac{1}{T} \sum_{t=1}^T \frac{|y_t - \hat{y}_t|}{(|y_t| + |\hat{y}_t|)/2 + \epsilon},
\end{equation}
where $y_t$ and $\hat{y}_t$ are the ground-truth and predicted values at time step $t$, $T$ is the total number of evaluation steps, and $\epsilon$ ensures numerical stability.
For sparse events such as Alerts and Upgrades, we report \emph{F1} on active frames.

\paragraph{Development.}
This dimension focuses on ongoing tasks and their progress. We calculate \emph{Queue F1} for task prediction accuracy, and \emph{Progress MAE} for temporal progress estimation.

\paragraph{Micro-Entity.}
We evaluate unit existence and attribute accuracy using a hybrid matching strategy between predicted units $p_j$ and ground-truth units $g_i$. A pair is counted as a true positive (TP) if either (i) \emph{ID-anchored}: $id(p_j)=id(g_i)$, or (ii) \emph{spatial-anchored}: $type(p_j)=type(g_i)$ and $\lVert pos(p_j)-pos(g_i)\rVert_2 \le \delta$. Unmatched predictions are false positives (FP), and unmatched ground-truth units are false negatives (FN), from which we compute \emph{Precision}, \emph{Recall}, and \emph{F1}. For matched pairs, we report attribute \emph{MAE} on health and energy.

\paragraph{Macro-Situation.}
The macro situation can be characterized by the spatial distribution of entities from both players. To measure the difference between predicted and ground-truth spatial distributions, inspired by optimal transport~\cite{OptimalTransport}, we introduce the \textbf{Augmented Wasserstein Distance (AWD)}. It calculates the minimum cost to transform the predicted distribution to the ground-truth distribution while penalizing unmatched entities.
Given the ground-truth set $G=\{g_i\}_{i=1}^M$ and the predicted set $P=\{p_j\}_{j=1}^N$, we formulate a linear sum assignment problem with an augmented cost matrix:
\begin{equation}
    \mathbf{C} = \begin{bmatrix}  \mathbf{D}_{match} & \mathbf{D}_{miss} \\  \mathbf{D}_{halluc} & \mathbf{0}  \end{bmatrix} \in\mathbb{R}^{(M+N)\times(N+M)},
\end{equation}
where $\mathbf{D}_{\text{match}}\in\mathbb{R}^{M\times N}$ contains pairwise Euclidean distance between predicted and ground-truth entities, with cross-type pairs assigned infinite cost. $\mathbf{D}_{\text{miss}}\in\mathbb{R}^{M\times M}$ and $\mathbf{D}_{\text{halluc}}\in\mathbb{R}^{N\times N}$ are diagonal matrices whose entries are set to a penalty $\lambda$ to penalize unmatched units.

We solve the assignment problem using the Hungarian algorithm~\cite{Hungarianmethod} to obtain the minimum total cost $\mathcal{L}_{total}$, and define the final metric:
\begin{equation}
    \text{AWD} = \frac{\mathcal{L}_{total}}{M + N}.
\end{equation}
Lower AWD indicates higher consistency with the ground-truth macro situation.

\subsection{World-Model-Augmented Decision System}
\begin{figure}[tb]
    \centering
    \includegraphics[width=\linewidth]{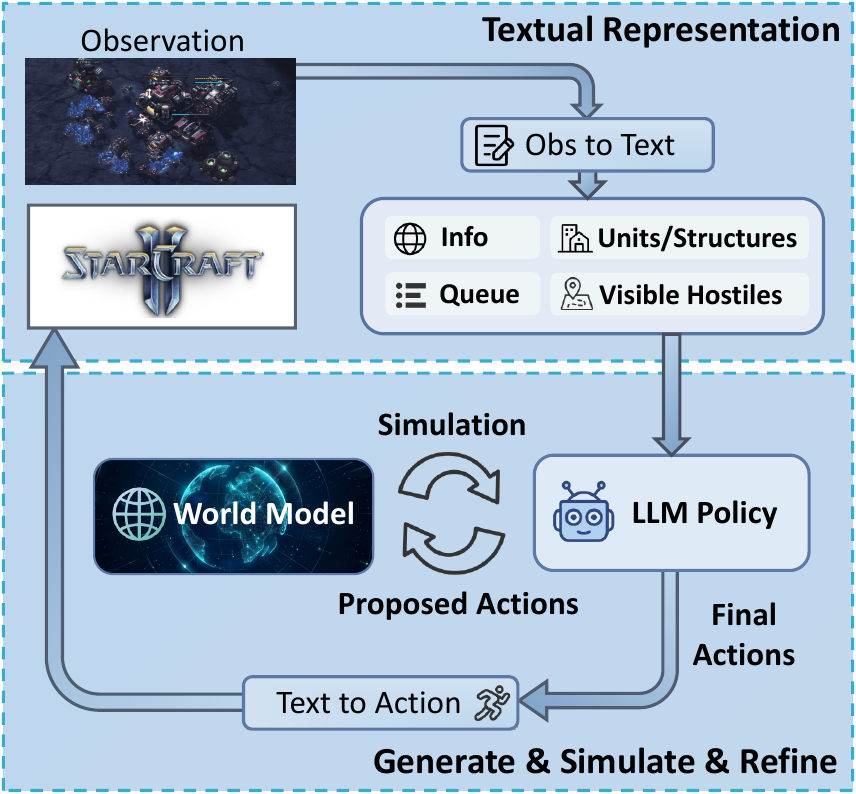} 
    \caption{Framework of our StarWM-Agent. The agent interacts with the SC2 engine via a textual interface and follows a \emph{Generate--Simulate--Refine} decision loop: the policy first generates an initial action proposal from the current observation, the world model predicts the short-horizon future observation, and the policy then makes the final decision conditioned on the predicted future.}
    \label{fig:starwm-agent}
\end{figure}

To study the decision utility of the learned world model, we develop \textbf{StarWM-Agent}, a world-model-augmented decision system that follows a lightweight \emph{Generate--Simulate--Refine} pipeline for foresight-driven policy refinement, as shown in Figure~\ref{fig:starwm-agent}. Given the current observation $o_t$, the policy $\pi_\theta$ first generates an initial action proposal
\begin{equation}
a_{\text{init}} = \pi_\theta(o_t),
\end{equation}
which is then simulated by the world model $\mathcal{M}_\phi$ to predict a short-horizon future observation
\begin{equation}
\hat{o}_{t+\tau} = \mathcal{M}_\phi(o_t, a_{\text{init}}).
\end{equation}
We then construct an augmented context
\begin{equation}
c_t = [o_t; a_{\text{init}}; \hat{o}_{t+\tau}],
\end{equation}
and obtain the final decision by a second policy call
\begin{equation}
a_{\text{final}} = \pi_\theta(c_t).
\end{equation}

This design aims to \textbf{balance decision performance and inference cost}: instead of performing expensive search over the large SC2 action space, the world model introduces external foresight into the policy's decision process through short-horizon simulation, and the refinement step allows the policy to revise the initial action proposal or keep it unchanged.

By incorporating a world model, StarWM-Agent obtains \textbf{two dimensions of decision enhancement}. At the macro-management level, StarWM extends the agent's effective decision horizon by forecasting resource flow, supply, and task progress, enabling the policy to plan ahead for upcoming bottlenecks such as resource shortages or supply caps. At the micro-tactical level, StarWM serves as a lightweight combat-and-feasibility simulator, supporting tactical risk assessment before execution, such as avoiding unfavorable engagements.

\section{Experiment}
Our experiments are designed around two research questions. In Section~\ref{subsection:offline_evaluation_results}, we conduct offline evaluations to answer \textbf{RQ1 (Learnability):} Can a trajectory-trained, action-conditioned world model capture the hybrid dynamics of SC2? In Section~\ref{subsection:online_testing_results}, we conduct online evaluation to answer \textbf{RQ2 (Decision Utility):} Does integrating a learned world model into the decision loop improve overall decision-making performance?


\begin{table*}[tb]
    \centering
    \resizebox{\linewidth}{!}{
    \begin{tabular}{lcccc}
        \toprule
        Method & 
        \begin{tabular}[c]{@{}c@{}}Economy\\ (Minerals / Gas SMAPE$\downarrow$)\end{tabular} & 
        \begin{tabular}[c]{@{}c@{}}Development\\ (Queue F1$\uparrow$ / Progress MAE\%$\downarrow$)\end{tabular} & 
        \begin{tabular}[c]{@{}c@{}}Micro-Entity\\ (Self / Enemy HP MAE\%$\downarrow$)\end{tabular} & 
        \begin{tabular}[c]{@{}c@{}}Macro-Situation\\ (Self / Enemy AWD$\downarrow$)\end{tabular} \\
        \midrule
        Static Bias & 0.49 / 0.25 & 0.80 / 24.81 & 4.81 / 9.15          & 8.37 / \textbf{16.13} \\
        Qwen3-8B    & 0.49 / 0.27 & 0.73 / 27.07 & 5.62 / 9.01          & 10.87 / 17.76 \\
        Qwen3-32B   & 0.48 / 0.26 & 0.70 / 24.27 & 5.11 / 8.47          & 9.79 / 17.78 \\
        \textbf{StarWM} & \textbf{0.19 / 0.09} & \textbf{0.92 / 0.43} & \textbf{4.15 / 7.90} & \textbf{3.46} / 18.09 \\
        \bottomrule
    \end{tabular}
    }
    \caption{Offline evaluation results. MAE for Progress and Health Percentage (HP) is computed as absolute difference in percentage. Our StarWM significantly outperforms baselines in predicting economy values, development progress, unit health, and self-side macro-situation. Detailed results are provided in the Appendix~\ref{appendix:Detailed_Offline_Results}.}
    \label{tab:offline_results}
\end{table*}

\subsection{Experimental Setup}\label{4-1Setup}
Our main experiments use Terran-versus-Terran games on the Flat64 map as a controlled SC2 evaluation setting for action-conditioned dynamics modeling and world-model-augmented decision-making. This setting retains SC2's large state-action space, partial observability, and hybrid dynamics, while enabling consistent offline and online comparisons across methods.

\paragraph{SC2-Dynamics-50K Dataset.}
Trajectories are collected by running a rule-based scripted agent against the SC2 built-in AI at Harder (LV6) and VeryHard (LV7), where both the scripted agent and the built-in AI use \textbf{randomized strategy styles}. 
We perform an 8:1:1 trajectory-level train/validation/test split before sample construction, and then use a 5-second prediction horizon (i.e., $\tau=5$) with a sliding-window step of 1 second, yielding 50,407 training samples, 6,774 validation samples, and 6,579 test samples. Detailed dataset construction and diversity analysis are provided in Appendix~\ref{appendix:The SC2-Dynamics-50K Dataset}, while a full sample is presented in Appendix~\ref{appendix:Example of SC2-Dynamics-50K}.

\paragraph{World Model Training.}
We use Qwen3-8B \cite{qwen3technicalreport} as the backbone of StarWM and perform supervised fine-tuning (SFT) with LLaMA-Factory~\cite{zheng2024llamafactory}. We adopt LoRA~\cite{Hu2021lora} with rank 8, a learning rate of 5$\times$$10^{-5}$, and train for 10 epochs on 8$\times$H100 GPUs.

\paragraph{Offline Evaluation Settings.}
We compare StarWM against three baselines: Static Bias (copying the input observation as the prediction), Qwen3-8B (zero-shot), and Qwen3-32B (zero-shot). Detailed settings are provided in Appendix~\ref{appendix:Offline Evaluation Settings.}.

\paragraph{Online Evaluation Settings.}
Online baselines include zero-shot Qwen3-8B/Qwen3-32B policies and a Chain-of-Summarization (CoS) baseline adapted from TextStarCraft II~\cite{textstarcraft2}. Matches are conducted against the SC2 built-in AI using randomized strategy styles. We report multiple online metrics, including win rate, supply block rate, resource conversion rate, kill-loss ratio, and valid action rate. Detailed online implementation and metric definitions are provided in Appendix~\ref{appendix:Online Testing Settings.}.

\subsection{Offline Evaluation of StarWM}\label{subsection:offline_evaluation_results}

\paragraph{Main Results.}As shown in Table~\ref{tab:offline_results}, StarWM yields the strongest performance across most metrics.
(1) \textbf{Economy:} For resource forecasting, StarWM achieves SMAPE of 0.19 / 0.09 for minerals and gas, respectively, significantly outperforming the zero-shot 32B baseline (0.48 / 0.26), with reductions of 60\% and 65\%.
(2) \textbf{Development:} StarWM reaches a Queue F1 of 0.92, while Progress MAE drops to only 0.43\%, whereas all baselines exceed 24\%. 
These results suggest that StarWM captures both task progression and macro-management logic.
(3) \textbf{Micro-level attributes:} Compared to zero-shot 32B's 5.11\% / 8.47\%, StarWM reduces the HP MAE for self/enemy units to 4.15\% / 7.90\%, and also significantly outperforms the Static Bias baseline. This indicates that StarWM effectively models combat attrition dynamics, enabling the simulation of health degradation under combat interactions.
(4) \textbf{Macro-Situation:} For the self-side entity distribution (including units and structures), compared to 8.37 for Static Bias and 9.79 for the zero-shot 32B baseline, StarWM reduces the AWD to 3.46, an improvement of nearly 60\%, indicating more accurate action-conditioned kinematics prediction.

\paragraph{Limits of Zero-Shot LLMs as World Models.}Table~\ref{tab:offline_results} shows that zero-shot LLMs (both 8B and 32B) fail to consistently outperform the simple Static Bias across most metrics. This weakness is not solely caused by invalid output formats: the parsing success rates of zero-shot 8B, 32B, and StarWM are 96.38\%, 97.54\%, and 99.44\%, respectively. Since unparsable outputs are excluded from metric computation and the numbers of parseable samples are comparable across methods, the remaining gap suggests that generic pre-trained language models lack prior knowledge of SC2's specific physical laws, and therefore cannot directly function as accurate forward dynamics models without task-specific adaptation. 


\paragraph{Boundary under Partial Observability.}
For enemy-side macro-situation, both StarWM and zero-shot LLMs perform slightly worse than Static Bias (e.g., AWD 18.09 vs. 16.13), highlighting a key boundary of single-frame prediction under Fog of War. Under partial observability, StarWM attempts to predict plausible enemy-side evolution (e.g., scouting or regrouping) conditioned on the current observation, but such predictions are inherently under-determined without sufficient temporal evidence. To probe whether temporal information helps, we further train a history-augmented StarWM using $(o_{t-2}, o_{t-1}, o_t, a_{t:t+\tau})$ to predict $o_{t+\tau}$. As shown in Table~\ref{tab:history_augmented_starwm}, temporal history improves the overall prediction performance. Enemy-side AWD also improves slightly, but remains challenging. This suggests that history helps, while more accurate prediction under partial observability points to the need for stronger temporal memory and belief-state modeling, which we leave for future work.

\paragraph{Case Study.}
Figure~\ref{fig:case_study} provides an intuitive example. The spatial distribution of self units predicted by StarWM (hollow circles) closely matches the ground truth (filled circles), preserving the army formation, whereas zero-shot LLM predictions appear more scattered and less coherent. Additional case studies and temporal analysis of macro-situation consistency are provided in Appendix~\ref{appendix:Temporal_and_Qualitative Analyses}.

\paragraph{Cross-map and Out-of-domain Opponent Probe.}
To probe the generalization of StarWM, we further evaluate it without retraining on the unseen Flat128 map against the built-in AI at Hard (LV5). As shown in Table~\ref{tab:cross_map_probe}, despite the distribution shift, StarWM remains strongest on most economy, development, and self-side metrics, with enemy AWD as the main exception. This suggests that StarWM learns dynamics that are not merely tied to the original training distribution, but generalize to some extent across map and opponent shifts. Details are provided in Appendix~\ref{appendix:Cross-Map and Out-of-Domain Opponent Probe}.

\paragraph{Action-Conditioned Prediction Probe.}
We further fix an early-game observation and evaluate StarWM under different single-action inputs. Figure~\ref{fig:action_conditioned_case_study} shows that the resulting future observations are distinct and logically consistent with the input actions, qualitatively illustrating that StarWM learns action-conditioned dynamics rather than merely reproducing surface-level textual patterns.

\paragraph{Prediction Horizon Analysis.}We further explore a longer prediction horizon by training StarWM with $\tau=10$ under the same training-data scale as the main setting. As shown in Table~\ref{tab:prediction_horizon}, although StarWM's prediction performance at $\tau=10$ degrades compared with that at $\tau=5$, it does not completely collapse and remains substantially better than Static Bias on most economy, development, and self-side metrics. Details are provided in Appendix~\ref{appendix:Prediction Horizon Analysis.}.

\subsection{Online Evaluation of StarWM-Agent}\label{subsection:online_testing_results}

\begin{table*}[t]
    \centering
    \resizebox{\textwidth}{!}{
    \begin{tabular}{lccccc}
        \toprule
        Setting & Win Rate(\%)$\uparrow$ & Supply Block Rate(\%)$\downarrow$ & Resource Conversion Rate(\%)$\uparrow$ & Kill-Loss Ratio(\%)$\uparrow$ & Valid Action Rate(\%)$\uparrow$ \\
        \midrule
        \multicolumn{6}{l}{\textbf{LV5 (Hard)}}\\
        \midrule
        Qwen3-8B      & 0.0\%  & $63.58 \pm 13.54$ & $29.09 \pm 19.14$ & 15.22\% & $17.89 \pm 12.25$ \\
        Qwen3-32B     & 20.0\% & $25.45 \pm 17.47$ & $52.32 \pm 23.03$ & 62.42\% & $41.31 \pm 20.81$ \\
        \rowcolor{oursbg} StarWM-Agent (8B)   & 10.0\% & $5.42 \pm 4.97$   & $84.20 \pm 14.21$ & 50.44\% & $86.29 \pm 12.88$ \\
        \rowcolor{oursbg} StarWM-Agent (32B)  & 50.0\% & $6.09 \pm 4.61$   & $81.11 \pm 7.56$  & 89.87\% & $85.57 \pm 11.01$ \\
        \midrule
        \multicolumn{6}{l}{\textbf{LV6 (Harder)}}\\
        \midrule
        Qwen3-8B      & 5.0\%  & $58.31 \pm 21.15$ & $36.12 \pm 21.36$ & 6.25\%  & $28.00 \pm 21.64$ \\
        Qwen3-32B     & 25.0\% & $21.41 \pm 21.29$ & $58.81 \pm 21.34$ & 27.42\% & $55.49 \pm 25.58$ \\
        \rowcolor{oursbg} StarWM-Agent (8B)   & 10.0\% & $10.57 \pm 9.28$  & $78.77 \pm 8.82$  & 28.12\% & $82.37 \pm 10.23$ \\
        \rowcolor{oursbg} StarWM-Agent (32B)  & 40.0\% & $5.93 \pm 5.20$   & $78.67 \pm 8.66$  & 41.30\% & $84.00 \pm 11.22$ \\
        \midrule
        \multicolumn{6}{l}{\textbf{LV7 (VeryHard)}}\\
        \midrule
        Qwen3-8B      & 0.0\%  & $58.74 \pm 19.99$ & $32.78 \pm 21.61$ & 12.30\% & $25.29 \pm 20.86$ \\
        Qwen3-32B     & 20.0\% & $16.39 \pm 21.25$ & $55.74 \pm 20.72$ & 29.26\% & $61.64 \pm 26.31$ \\
        \rowcolor{oursbg} StarWM-Agent (8B)   & 20.0\% & $5.51 \pm 5.21$   & $82.89 \pm 8.96$  & 20.00\% & $82.13 \pm 11.91$ \\
        \rowcolor{oursbg} StarWM-Agent (32B)  & 50.0\% & $5.39 \pm 4.90$   & $76.27 \pm 16.14$ & 50.51\% & $81.99 \pm 22.50$ \\
        \bottomrule
    \end{tabular}
    }
    \caption{Online evaluation against SC2's built-in AI at different difficulty levels. StarWM-Agent (8B/32B) denotes our world-model-augmented decision system using zero-shot Qwen3-8B/32B as the policy model. Each setting is evaluated over 20 matches. Note that LV5 is an out-of-distribution (OOD) opponent, as StarWM was trained only on LV6 and LV7 trajectories.}
    \label{tab:online_metrics}
\end{table*}

\begin{table*}[t]
\centering
\resizebox{\textwidth}{!}{
\begin{tabular}{llcccccc}
\toprule
Setting & Pipeline & Win$\uparrow$ & SBR$\downarrow$ & RCR$\uparrow$ & KLR$\uparrow$ & VAR$\uparrow$ & Latency (s)$\downarrow$ \\
\midrule
Policy-only & Generate & 0.0 & $58.74 \pm 19.99$ & $32.78 \pm 21.61$ & 12.30 & $25.29 \pm 20.86$ & $1.96 \pm 0.62$ \\
CoS & CoS + Generate & 10.0 & $31.00 \pm 21.51$ & $56.07 \pm 17.03$ & 11.34 & $54.37 \pm 16.16$ & $6.34 \pm 0.81$ \\
StarWM-Agent & Generate + Simulate + Refine & 20.0 & $\textbf{5.51} \pm \textbf{5.21}$ & $82.89 \pm 8.96$ & 20.00 & $82.13 \pm 11.91$ & $11.61 \pm 3.78$ \\
CoS + StarWM-Agent & CoS + Generate + Simulate + Refine & \textbf{35.0} & $6.13 \pm 7.48$ & $\textbf{86.92} \pm \textbf{10.00}$ & \textbf{29.40} & $\textbf{83.16} \pm \textbf{15.02}$ & $14.92 \pm 3.25$ \\
\bottomrule
\end{tabular}
}
\caption{Comparison with Chain-of-Summarization (CoS) against the SC2 built-in AI at LV7 (VeryHard), using zero-shot Qwen3-8B as the policy model. Metrics except latency are percentages (\%); latency denotes the average per-decision inference time. The CoS baseline is adapted from TextStarCraft II~\cite{textstarcraft2}.}
\label{tab:cos_latency}
\end{table*}

\paragraph{Overall Performance.}
Table~\ref{tab:online_metrics} shows that our StarWM-Agent achieves the best overall performance across all difficulty levels. Compared to the baselines, StarWM-Agent (8B) improves win rates by 10\% / 5\% / 20\% against LV5 / LV6 / LV7, respectively, while StarWM-Agent (32B) achieves larger gains of 30\% / 15\% / 30\%. Notably, although StarWM is trained only on LV6 and LV7 trajectories, StarWM-Agent still achieves substantial improvements against the unseen LV5 opponent.

\paragraph{Macro and Tactical Effects.}
Beyond win rate, StarWM-Agent also improves decision quality in both macro-management and micro-tactical dimensions. At the macro-management level, it reduces Supply Block Rate (SBR) by approximately 53\% for the 8B policy and 15\% for the 32B policy, while improving Resource Conversion Rate (RCR) by 49\% and 23\%, respectively. These gains suggest that world-model lookahead helps the policy anticipate upcoming supply bottlenecks, resource shortages, and idle production facilities, leading to more proactive macro-management decisions. At the micro-tactical level, Kill-Loss Ratio (KLR) improves by around 21\% for both policies, suggesting that world-model simulation helps the policy assess tactical risks and make better combat decisions.

\paragraph{Complementarity with CoS.}
To position StarWM-Agent against existing SC2 LLM-agent mechanisms, we further compare it with a Chain-of-Summarization (CoS) baseline adapted from TextStarCraft II~\cite{textstarcraft2}, a representative existing method for compressing observation histories before action generation. As shown in Table~\ref{tab:cos_latency}, CoS improves win rate from 0\% to 10\%, while StarWM-Agent reaches 20\%; combining CoS with StarWM-Agent further improves win rate to 35\% and achieves the best RCR, KLR, and VAR. This suggests that our methods are complementary to CoS: CoS compresses observation histories, whereas StarWM provides action-conditioned forward simulation. The average decision latency also reveals a clear performance--cost trade-off under the common LLM-SC2 protocol that pauses the engine at decision frames.


\begin{table}[t]
\centering
\resizebox{\columnwidth}{!}{
\begin{tabular}{lcccccc}
\toprule
Setting & Win$\uparrow$ & SBR$\downarrow$ & RCR$\uparrow$ & KLR$\uparrow$ & VAR$\uparrow$ & ARR \\
\midrule
Generate (Policy-only) & 0.0 & 58.74 & 32.78 & 12.30 & 25.29 & -- \\
+ Refine & 5.0 & 9.19 & 76.44 & 13.97 & 78.04 & 34.64 \\
+ Zero-shot WM Simulate & 10.0 & 9.98 & 72.80 & 14.80 & 78.44 & 26.66 \\
+ StarWM Simulate & \textbf{20.0} & \textbf{5.51} & \textbf{82.89} & \textbf{20.00} & \textbf{82.13} & 32.33 \\
\bottomrule
\end{tabular}
}
\caption{Ablation study against LV7 (VeryHard), using zero-shot Qwen3-8B as the policy model. Metrics are percentages (\%); ARR denotes Action Revision Rate.}
\label{tab:ablation_8b}
\end{table}

\paragraph{Ablation Study.}
To isolate the source of StarWM-Agent's performance gains, we conduct an internal ablation against LV7 (VeryHard) using zero-shot Qwen3-8B as the policy model (Table~\ref{tab:ablation_8b}). Specifically, we compare four configurations with progressively added components: (1) \emph{Generate}, where the policy directly outputs actions; (2) \emph{Generate + Refine}, which enables self-reflection; (3) \emph{Generate + Zero-shot WM Simulate + Refine}, where a zero-shot Qwen3-8B serves as a world model for forward simulation with the same prompt template as StarWM; and (4) \emph{Generate + StarWM Simulate + Refine} (i.e., StarWM-Agent), which incorporates the trained world model.

Compared with direct generation, self-reflection substantially improves macro-management metrics, reducing Supply Block Rate from 58.74\% to 9.19\% and increasing Resource Conversion Rate from 32.78\% to 76.44\%, with a modest win-rate gain from 0\% to 5\%. Adding zero-shot world-model simulation further improves win rate to 10\% and Kill-Loss Ratio to 14.80\%, suggesting that external predictive signals provide value beyond self-reflection alone. Finally, incorporating the trained StarWM achieves the best overall performance, increasing win rate to 20\% and improving all key metrics. These results suggest that StarWM-Agent's gains come not only from additional inference-time computation, but also from more accurate action-conditioned simulation.

\paragraph{Action Revision Analysis.}
Under the main settings (Table~\ref{tab:online_metrics}), the Action Revision Rates of StarWM-Agent (8B) and StarWM-Agent (32B) are 32.74\% and 19.45\%, respectively, aggregated across the three opponents.
As shown in Figure~\ref{fig:action_analysis_32b}, \emph{Build Supply Depot} is the most frequent added action for StarWM-Agent (32B), consistent with its improved macro-management. The case in Figure~\ref{fig:online_casestudy} further illustrates how world-model simulation helps refine a suboptimal initial action.



\section{Conclusion}
We present StarWM, a player-view, action-conditioned textual world model for SC2, together with StarWM-Agent, a world-model-augmented decision system for foresight-driven policy refinement. Offline evaluation shows the \textbf{learnability} of key SC2 dynamics from trajectory data, while online evaluation demonstrates the \textbf{decision utility} of integrating the learned world model into the policy loop. Our analyses also characterize the boundaries of the current formulation. We believe this work provides useful insights for the community and motivates future research on world models for complex partially observable competitive environments.

\section*{Limitations}
\paragraph{Temporal Modeling under Partial Observability.}
As discussed in Section~\ref{subsection:offline_evaluation_results}, our offline analysis identifies a boundary of the current single-frame prediction formulation: future enemy-side states are not fully determined by the current observation under Fog of War. Our history-augmented experiment further shows that temporal history can improve prediction, suggesting that additional temporal evidence is useful. More accurate prediction under partial observability may require stronger temporal memory and belief-state modeling, which we leave for future work.


\paragraph{Controlled Evaluation Scope.}
Our main experiments are conducted in a controlled SC2 evaluation setting against built-in AI opponents with randomized strategy styles, enabling consistent evaluation of action-conditioned dynamics modeling and decision utility. Although we include preliminary probes beyond the main setting, such as additional-map and out-of-domain-opponent probes, these probes are not intended as a comprehensive evaluation across the full diversity of SC2 maps and race matchups. Extending StarWM to more diverse maps and asymmetric race matchups remains an important direction for future work.


\paragraph{Inference Cost and Real-Time Deployment.}
Existing LLM-based SC2 agents mainly prioritize improving policy decision quality over satisfying strict real-time APM constraints. In online matches, they typically pause the game engine at decision frames, wait for agent inference, and then resume execution with the generated actions. Our evaluation follows this common protocol, while we additionally report latency analysis to make the performance--cost trade-off explicit. For stricter real-time deployment, future work may reduce inference cost by predicting only state deltas instead of full future observations, distilling StarWM into a smaller model, or improving serving efficiency.

\paragraph{Textual Representation Fidelity.}
The textual representation is a task-oriented compression of engine-side structured information rather than a lossless encoding. The observation-to-text operator $\Phi(\cdot)$ preserves information most relevant to short-horizon dynamics modeling, while abstracting away some fine-grained entity attributes. Future work may further improve representation fidelity by incorporating richer structured signals or multimodal state representations.


\section*{Acknowledgments}
This work is supported by the New Generation Artificial Intelligence--National Science and Technology Major Project (No.2025ZD0124203).

\bibliography{custom}

@inproceedings{yao2023react,
  title = {{ReAct}: Synergizing Reasoning and Acting in Language Models},
  author = {Yao, Shunyu and Zhao, Jeffrey and Yu, Dian and Du, Nan and Shafran, Izhak and Narasimhan, Karthik and Cao, Yuan},
  booktitle = {International Conference on Learning Representations (ICLR) },
  year = {2023},
  html = {https://arxiv.org/abs/2210.03629},
}

@inproceedings{schick2023toolformer,
  title={Toolformer: language models can teach themselves to use tools},
  author={Schick, Timo and Dwivedi-Yu, Jane and Dess{\'\i}, Roberto and Raileanu, Roberta and Lomeli, Maria and Hambro, Eric and Zettlemoyer, Luke and Cancedda, Nicola and Scialom, Thomas},
  booktitle={Proceedings of the 37th International Conference on Neural Information Processing Systems},
  pages={68539--68551},
  year={2023}
}

@article{wang2023voyager,
title={Voyager: An Open-Ended Embodied Agent with Large Language Models},
author={Guanzhi Wang and Yuqi Xie and Yunfan Jiang and Ajay Mandlekar and Chaowei Xiao and Yuke Zhu and Linxi Fan and Anima Anandkumar},
journal={Transactions on Machine Learning Research},
issn={2835-8856},
year={2024},
url={https://openreview.net/forum?id=ehfRiF0R3a},
note={}
}

@inproceedings{OptimalTransport,
  title={Optimal Transport: Old and New},
  author={C{\'e}dric Villani},
  year={2008},
  url={https://api.semanticscholar.org/CorpusID:118347220}
}

@article{Hungarianmethod,
  title={The Hungarian method for the assignment problem},
  author={Harold W. Kuhn},
  journal={Naval Research Logistics (NRL)},
  year={1955},
  volume={52},
  url={https://api.semanticscholar.org/CorpusID:9426884}
}

@article{muzero,
  title = {Mastering {{Atari}}, {{Go}}, Chess and Shogi by Planning with a Learned Model},
  author = {Schrittwieser, Julian and Antonoglou, Ioannis and Hubert, Thomas and Simonyan, Karen and Sifre, Laurent and Schmitt, Simon and Guez, Arthur and Lockhart, Edward and Hassabis, Demis and Graepel, Thore and Lillicrap, Timothy and Silver, David},
  year = 2020,
  month = dec,
  journal = {Nature},
  volume = {588},
  number = {7839},
  pages = {604--609},
  issn = {1476-4687},
  doi = {10.1038/s41586-020-03051-4}
}

@article{dreamerv3,
  title = {Mastering Diverse Control Tasks through World Models},
  author = {Hafner, Danijar and Pasukonis, Jurgis and Ba, Jimmy and Lillicrap, Timothy},
  year = 2025,
  month = apr,
  journal = {Nature},
  volume = {640},
  number = {8059},
  pages = {647--653},
  issn = {1476-4687},
  doi = {10.1038/s41586-025-08744-2}
}

@inproceedings{mbpo,
 author = {Janner, Michael and Fu, Justin and Zhang, Marvin and Levine, Sergey},
 booktitle = {Advances in Neural Information Processing Systems},
 editor = {H. Wallach and H. Larochelle and A. Beygelzimer and F. d\textquotesingle Alch\'{e}-Buc and E. Fox and R. Garnett},
 pages = {},
 publisher = {Curran Associates, Inc.},
 title = {When to Trust Your Model: Model-Based Policy Optimization},
 url = {https://proceedings.neurips.cc/paper_files/paper/2019/file/5faf461eff3099671ad63c6f3f094f7f-Paper.pdf},
 volume = {32},
 year = {2019}
}

@InProceedings{drivewm,
    author    = {Wang, Yuqi and He, Jiawei and Fan, Lue and Li, Hongxin and Chen, Yuntao and Zhang, Zhaoxiang},
    title     = {Driving into the Future: Multiview Visual Forecasting and Planning with World Model for Autonomous Driving},
    booktitle = {Proceedings of the IEEE/CVF Conference on Computer Vision and Pattern Recognition (CVPR)},
    month     = {June},
    year      = {2024},
    pages     = {14749-14759}
}

@misc{GAIA2,
      title={GAIA-2: A Controllable Multi-View Generative World Model for Autonomous Driving}, 
      author={Lloyd Russell and Anthony Hu and Lorenzo Bertoni and George Fedoseev and Jamie Shotton and Elahe Arani and Gianluca Corrado},
      year={2025},
      eprint={2503.20523},
      archivePrefix={arXiv},
      primaryClass={cs.CV},
      url={https://arxiv.org/abs/2503.20523}, 
}

@article{
webdreamer,
title={Is Your {LLM} Secretly a World Model of the Internet? Model-Based Planning for Web Agents},
author={Yu Gu and Kai Zhang and Yuting Ning and Boyuan Zheng and Boyu Gou and Tianci Xue and Cheng Chang and Sanjari Srivastava and Yanan Xie and Peng Qi and Huan Sun and Yu Su},
journal={Transactions on Machine Learning Research},
issn={2835-8856},
year={2025},
url={https://openreview.net/forum?id=c6l7yA0HSq},
note={}
}

@article{textstarcraft2,
  title={Large language models play starcraft ii: Benchmarks and a chain of summarization approach},
  author={Ma, Weiyu and Mi, Qirui and Zeng, Yongcheng and Yan, Xue and Lin, Runji and Wu, Yuqiao and Wang, Jun and Zhang, Haifeng},
  journal={Advances in Neural Information Processing Systems},
  volume={37},
  pages={133386--133442},
  year={2024}
}

@inproceedings{LLMPySC2SI,
title={{LLM}-Py{SC}2: Starcraft {II} learning environment for Large Language Models},
author={Zongyuan Li and Yanan Ni and Runnan Qi and Chang Lu and Lumin Jiang and Xu Xiaojie and Xiangbei Liu and Pengfei Li and Yunzheng Guo and Zhe Ma and Huanyu Li and wu hui and Guo Xian and Kuihua Huang and Xuebo Zhang},
booktitle={The Thirty-ninth Annual Conference on Neural Information Processing Systems},
year={2025},
url={https://openreview.net/forum?id=Xr73jEYG29}
}

@misc{SC2Arena,
      title={SC2Arena and StarEvolve: Benchmark and Self-Improvement Framework for LLMs in Complex Decision-Making Tasks}, 
      author={Pengbo Shen and Yaqing Wang and Ni Mu and Yao Luan and Runpeng Xie and Senhao Yang and Lexiang Wang and Hao Hu and Shuang Xu and Yiqin Yang and Bo Xu},
      year={2025},
      eprint={2508.10428},
      archivePrefix={arXiv},
      primaryClass={cs.LG},
      url={https://arxiv.org/abs/2508.10428}, 
}

@INPROCEEDINGS{CommCoT,
  author={Qi, Runnan and Quan, Yuming and Ni, Yanan and Li, Zongyuan and Xu, Xiaojie and Huang, Kuihua and Guo, Xian},
  booktitle={2025 IEEE 2nd International Conference on Electronics, Communications and Intelligent Science (ECIS)}, 
  title={Comm-CoT: Standardized Chain-of-Thought Communication Framework for Efficient LLM based Multi-Agent Decision-Making in Real-Time Strategy Games}, 
  year={2025},
  volume={},
  number={},
  pages={1-8},
  doi={10.1109/ECIS65594.2025.11087008}}

@misc{MASMP,
      title={Memory-Augmented State Machine Prompting: A Novel LLM Agent Framework for Real-Time Strategy Games}, 
      author={Runnan Qi and Yanan Ni and Lumin Jiang and Zongyuan Li and Kuihua Huang and Xian Guo},
      year={2025},
      eprint={2510.18395},
      archivePrefix={arXiv},
      primaryClass={cs.AI},
      url={https://arxiv.org/abs/2510.18395}, 
}

@article{buildmachineslikepeople,
  title={Building machines that learn and think like people},
  author={Lake, Brenden M and Ullman, Tomer D and Tenenbaum, Joshua B and Gershman, Samuel J},
  journal={Behavioral and brain sciences},
  volume={40},
  pages={e253},
  year={2017},
  publisher={Cambridge University Press}
}

@article{alphastar,
  title={Grandmaster level in StarCraft II using multi-agent reinforcement learning},
  author={Oriol Vinyals and Igor Babuschkin and Wojciech M. Czarnecki and Micha{\"e}l Mathieu and Andrew Joseph Dudzik and Junyoung Chung and David Choi and Richard Powell and Timo Ewalds and Petko Georgiev and Junhyuk Oh and Dan Horgan and Manuel Kroiss and Ivo Danihelka and Aja Huang and L. Sifre and Trevor Cai and John P. Agapiou and Max Jaderberg and Alexander Sasha Vezhnevets and R{\'e}mi Leblond and Tobias Pohlen and Valentin Dalibard and David Budden and Yury Sulsky and James Molloy and Tom Le Paine and Caglar Gulcehre and Ziyun Wang and Tobias Pfaff and Yuhuai Wu and Roman Ring and Dani Yogatama and Dario W{\"u}nsch and Katrina McKinney and Oliver Smith and Tom Schaul and Timothy P. Lillicrap and Koray Kavukcuoglu and Demis Hassabis and Chris Apps and David Silver},
  journal={Nature},
  year={2019},
  volume={575},
  pages={350 - 354},
  url={https://api.semanticscholar.org/CorpusID:204972004}
}

@inproceedings{starcraftdefogger,
  title = {Forward Modeling for Partial Observation Strategy Games - a {{StarCraft}} Defogger},
  booktitle = {Advances in Neural Information Processing Systems},
  author = {Synnaeve, Gabriel and Lin, Zeming and Gehring, Jonas and Gant, Dan and Mella, Vegard and Khalidov, Vasil and Carion, Nicolas and Usunier, Nicolas},
  editor = {Bengio, S. and Wallach, H. and Larochelle, H. and Grauman, K. and {Cesa-Bianchi}, N. and Garnett, R.},
  year = 2018,
  volume = {31},
  publisher = {Curran Associates, Inc.}
}

@inproceedings{hao2023reasoning,
    title = "Reasoning with Language Model is Planning with World Model",
    author = "Hao, Shibo  and
      Gu, Yi  and
      Ma, Haodi  and
      Hong, Joshua  and
      Wang, Zhen  and
      Wang, Daisy  and
      Hu, Zhiting",
    editor = "Bouamor, Houda  and
      Pino, Juan  and
      Bali, Kalika",
    booktitle = "Proceedings of the 2023 Conference on Empirical Methods in Natural Language Processing",
    month = dec,
    year = "2023",
    address = "Singapore",
    publisher = "Association for Computational Linguistics",
    url = "https://aclanthology.org/2023.emnlp-main.507/",
    doi = "10.18653/v1/2023.emnlp-main.507",
    pages = "8154--8173"
}

@inproceedings{zheng2024llamafactory,
  title={LlamaFactory: Unified Efficient Fine-Tuning of 100+ Language Models},
  author={Yaowei Zheng and Richong Zhang and Junhao Zhang and Yanhan Ye and Zheyan Luo and Zhangchi Feng and Yongqiang Ma},
  booktitle={Proceedings of the 62nd Annual Meeting of the Association for Computational Linguistics (Volume 3: System Demonstrations)},
  address={Bangkok, Thailand},
  publisher={Association for Computational Linguistics},
  year={2024},
  url={http://arxiv.org/abs/2403.13372}
}

@inproceedings{Hu2021lora,
title={Lo{RA}: Low-Rank Adaptation of Large Language Models},
author={Edward J Hu and Yelong Shen and Phillip Wallis and Zeyuan Allen-Zhu and Yuanzhi Li and Shean Wang and Lu Wang and Weizhu Chen},
booktitle={International Conference on Learning Representations},
year={2022},
url={https://openreview.net/forum?id=nZeVKeeFYf9}
}

@misc{qwen3technicalreport,
      title={Qwen3 Technical Report}, 
      author={Qwen Team},
      year={2025},
      eprint={2505.09388},
      archivePrefix={arXiv},
      primaryClass={cs.CL},
      url={https://arxiv.org/abs/2505.09388}, 
}

@inproceedings{bleu,
    title = "{B}leu: a Method for Automatic Evaluation of Machine Translation",
    author = "Papineni, Kishore  and
      Roukos, Salim  and
      Ward, Todd  and
      Zhu, Wei-Jing",
    editor = "Isabelle, Pierre  and
      Charniak, Eugene  and
      Lin, Dekang",
    booktitle = "Proceedings of the 40th Annual Meeting of the Association for Computational Linguistics",
    month = jul,
    year = "2002",
    address = "Philadelphia, Pennsylvania, USA",
    publisher = "Association for Computational Linguistics",
    url = "https://aclanthology.org/P02-1040/",
    doi = "10.3115/1073083.1073135",
    pages = "311--318"
}

@inproceedings{rouge,
    title = "{ROUGE}: A Package for Automatic Evaluation of Summaries",
    author = "Lin, Chin-Yew",
    booktitle = "Text Summarization Branches Out",
    month = jul,
    year = "2004",
    address = "Barcelona, Spain",
    publisher = "Association for Computational Linguistics",
    url = "https://aclanthology.org/W04-1013/",
    pages = "74--81"
}

\clearpage
\appendix

\section{Introduction to StarCraft II}
\label{appendix:sc2_intro}

\paragraph{Game Overview.}StarCraft~II (SC2) is a real-time strategy (RTS) game developed by Blizzard Entertainment and is widely regarded as a benchmark for complex sequential decision-making. In the standard competitive 1v1 setting, the game can be modeled as a two-player zero-sum partially observable stochastic game. Each player selects one of three asymmetric races---Terran, Protoss, or Zerg---with race-specific units, technologies, and strategic styles.

\paragraph{Complexity.}SC2 integrates both macro-management and micro-tactical control. At the macro level, players collect minerals and vespene gas, maintain supply capacity, expand bases, and advance through technology trees to unlock stronger units. At the micro-tactical level, players control unit movement, positioning, and engagement decisions during battles, requiring coordination among heterogeneous units under time pressure. The state and action spaces are high-dimensional and combinatorial, making long-horizon planning and execution particularly challenging.

\paragraph{Dynamics Characteristics.}SC2 dynamics are hybrid and heterogeneous. Resource quantities change according to collection rates and action costs. Task progress follows deterministic build-time rules. Unit positions evolve under kinematic constraints, while unit health is governed by interaction-based combat dynamics. Meanwhile, due to the Fog of War, opponent states and actions are highly unobservable and must be inferred from limited information. This mixture of deterministic rules, interaction-driven effects, and partial observability makes accurate modeling of SC2 dynamics particularly challenging.

\section{Details and Examples of Textual Representation}\label{appendix:Textual Representation}
\subsection{Textual Representation}
\paragraph{Textual Observation Representation.}We propose a textual observation representation $\Phi(\cdot)$, explicitly factorizing observation into five distinct semantic modules. This design aims to decompose the observation-level dynamics into a set of sub-dynamics. The observation $o_t$ is structured as:
\begin{enumerate}
    \item \textbf{Info:} Describes economy and status (Minerals, Gas, Collection Rate, Supply, Alerts, Upgrades). This module isolates numerical flow from spatial complexity.
    \item \textbf{Queue:} Records ongoing tasks (construction, production, upgrades) and their progress. This module focuses on deterministic temporal progression.
    \item \textbf{My Units:} Includes self units' IDs, positions, health percentage (HP), energy and status. This module focuses on kinematics and damage interactions.
    \item \textbf{My Structures:} Describes self static assets. This module separates structures from units to focus on state switching.
    \item \textbf{Visible Hostiles:} Includes visible enemy units, structures, and snapshot enemy structures under the fog of war. This module isolates uncertainty handling for partial observability.
\end{enumerate}

\paragraph{Textual Action Representation.}
The textual action representation $\Psi(\cdot)$ follows a three-part format:
\[
\texttt{subject\_type [id] - action - target}.
\]
It consists of the acting entity, the action name, and an optional target field. Depending on the subject and action type, the target field takes one of the following forms:

\begin{itemize}
    \item Entity target. For actions targeting a specific entity (e.g., repair or heal), the target is represented as
\[
\texttt{target\_type [id]}.
\]
    \item Spatial target. For actions targeting a location (e.g., move), the target is represented as coordinates.
    \item Empty target. For actions without an explicit target (e.g., production-related actions), the target field is left empty.
\end{itemize}

\begin{figure*}[tb]
    \centering
    \includegraphics[width=\linewidth]{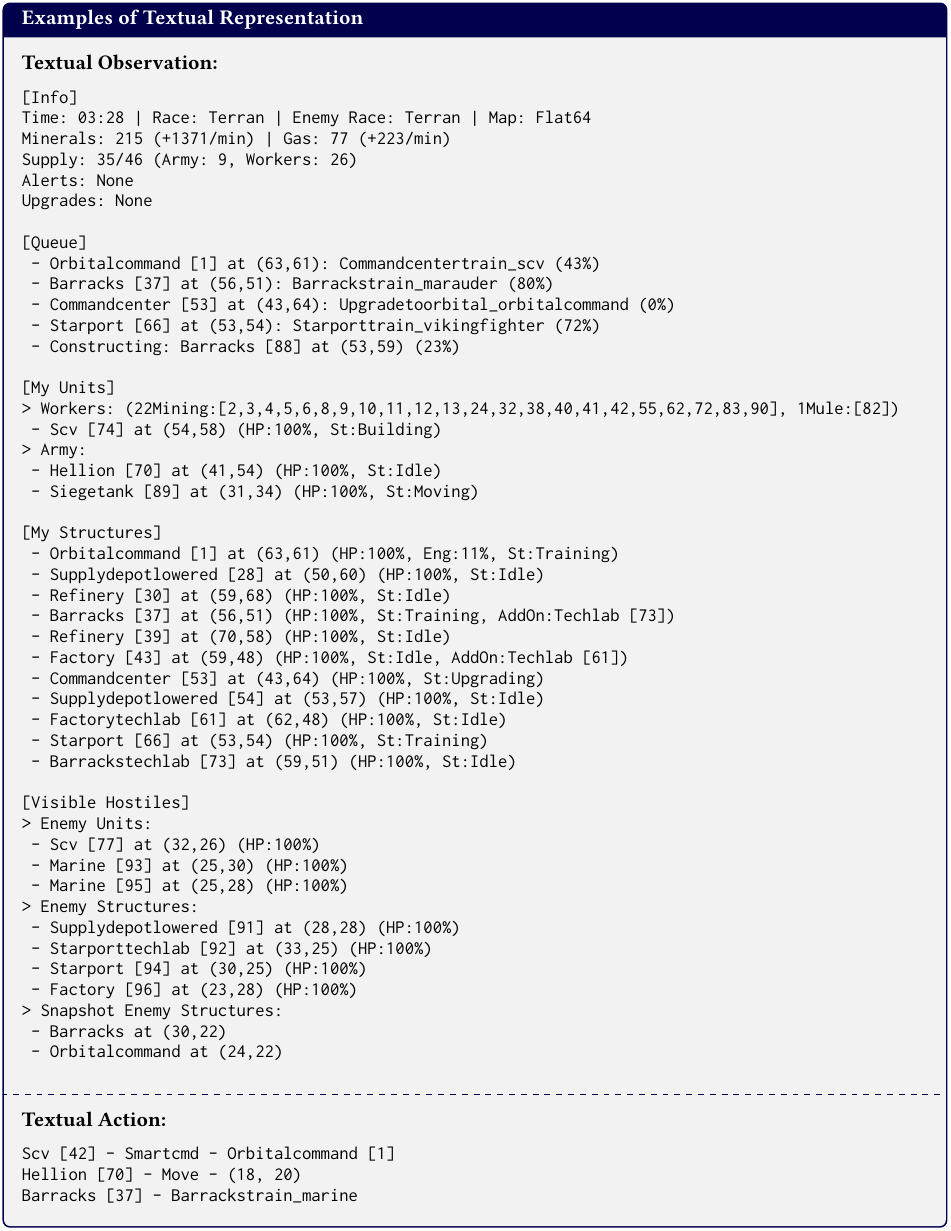} 
    \caption{An example of the textual observation and textual action.}
    \label{fig:Examples of Textual Representation}
\end{figure*}

\paragraph{Examples of Textual Representation.}
Figure~\ref{fig:Examples of Textual Representation} shows an example of the textual observation and textual action used in StarWM.

\subsection{Potential Information Loss}\label{appendix:Potential Information Loss}
The textual representation is a task-oriented compression of the engine-side structured information, rather than a lossless encoding. Specifically, the observation-to-text operator $\Phi(\cdot)$ is lossy: it preserves information most relevant to short-horizon dynamics modeling, while omitting certain entity-level attributes. In contrast, the action-to-text operator $\Psi(\cdot)$ is a near-bijective mapping between textual and executable engine actions, with the only loss arising from the quantization of floating-point coordinates into integers. Table~\ref{tab:attr_loss} summarizes representative retained and omitted attributes for entity-level attributes.

\begin{table}[htb]
\centering
\small
\setlength{\tabcolsep}{4pt}
\renewcommand{\arraystretch}{1.05}
\rowcolors{2}{gray!8}{white}
\begin{tabularx}{\columnwidth}{@{}p{0.4\columnwidth}cp{0.42\columnwidth}@{}}
\toprule
\rowcolor{white}
\textbf{Raw Engine Attr.} & \textbf{Used} & \textbf{Note} \\
\midrule
\texttt{type\_name} & \cmark &  \\
\texttt{tag} & \cmark & Convert to integer ID \\
\texttt{owner\_id} & \cmark & \\
\texttt{position} & \cmark & Quantize from float to integer \\
\texttt{health}, \texttt{health\_max} & \cmark & Represent as health percentage \\
\texttt{energy}, \texttt{energy\_max} & \cmark & Represent as energy percentage \\
\texttt{buffs} & \cmark &  \\
\texttt{build\_progress}, \texttt{orders} & \cmark & Group into \textit{Queue}\\
\texttt{add\_on\_tag} & \cmark &  \\
\texttt{is\_flying} & \cmark &  \\
\texttt{is\_snapshot}, \texttt{is\_visible} & \cmark &  \\
\texttt{attack\_upgrade\_level}, \texttt{armor\_upgrade\_level}, \texttt{shield\_upgrade\_level} & \cmark & Group into \textit{Upgrades} \\
\midrule
\texttt{shield}, \texttt{shield\_max} & \xmark & \\
\texttt{cloak}, \texttt{is\_burrowed}, \texttt{is\_powered}, \texttt{is\_hallucination}, \texttt{is\_active} & \xmark &  \\
\texttt{facing}, \texttt{radius} & \xmark & \\
\texttt{detect\_range}, \texttt{radar\_range} & \xmark & \\
\texttt{weapon\_cooldown} & \xmark & \\
\texttt{age}, \texttt{age\_in\_frames} & \xmark & \\
\bottomrule
\end{tabularx}
\caption{Entity-level attributes retained in or omitted from the textual observation. Retained attributes may still be compressed (e.g., coordinate quantization or percentage-based normalization).}
\label{tab:attr_loss}
\end{table}

\clearpage

\section{Details of Experimental Setup}
\subsection{SC2-Dynamics-50K Dataset}\label{appendix:The SC2-Dynamics-50K Dataset}
We build \emph{SC2-Dynamics-50K}, the first instruction-tuning dataset for SC2 dynamics prediction. The dataset is constructed from Terran vs. Terran (TvT) games on the Flat64 map, as this setting provides sufficient complexity for evaluating our core methodology.

\paragraph{Data Collection.}We collect trajectories by running a rule-based agent against SC2's built-in AI at Harder (LV6) and VeryHard (LV7), with 50 trajectories per level (100 trajectories in total). Both the scripted agent and the built-in AI use \textbf{randomized strategy styles}. Specifically, the built-in AI samples from Rush, Timing, Power, Macro, and Air styles, while the scripted agent uses 126 randomized opening styles and can adaptively switch strategies during a game according to the current situation. This design increases trajectory diversity and broadens coverage of the state-action distribution.

\paragraph{Data Processing.}We then split trajectories into train/validation/test with a ratio of 8:1:1. We set the prediction horizon to 5 seconds (i.e., $\tau=5$) with a sliding window step of 1 second. Using the textual representation, we convert trajectories into instruction-tuning triples of the form:
\begin{equation}
(o_t,\ a_{t:t+\tau}) \rightarrow o_{t+\tau}.
\end{equation}
This yields 50,407 training samples, 6,774 validation samples, and 6,579 test samples. 

\begin{figure}[htb]
    \centering
    \includegraphics[width=\linewidth]{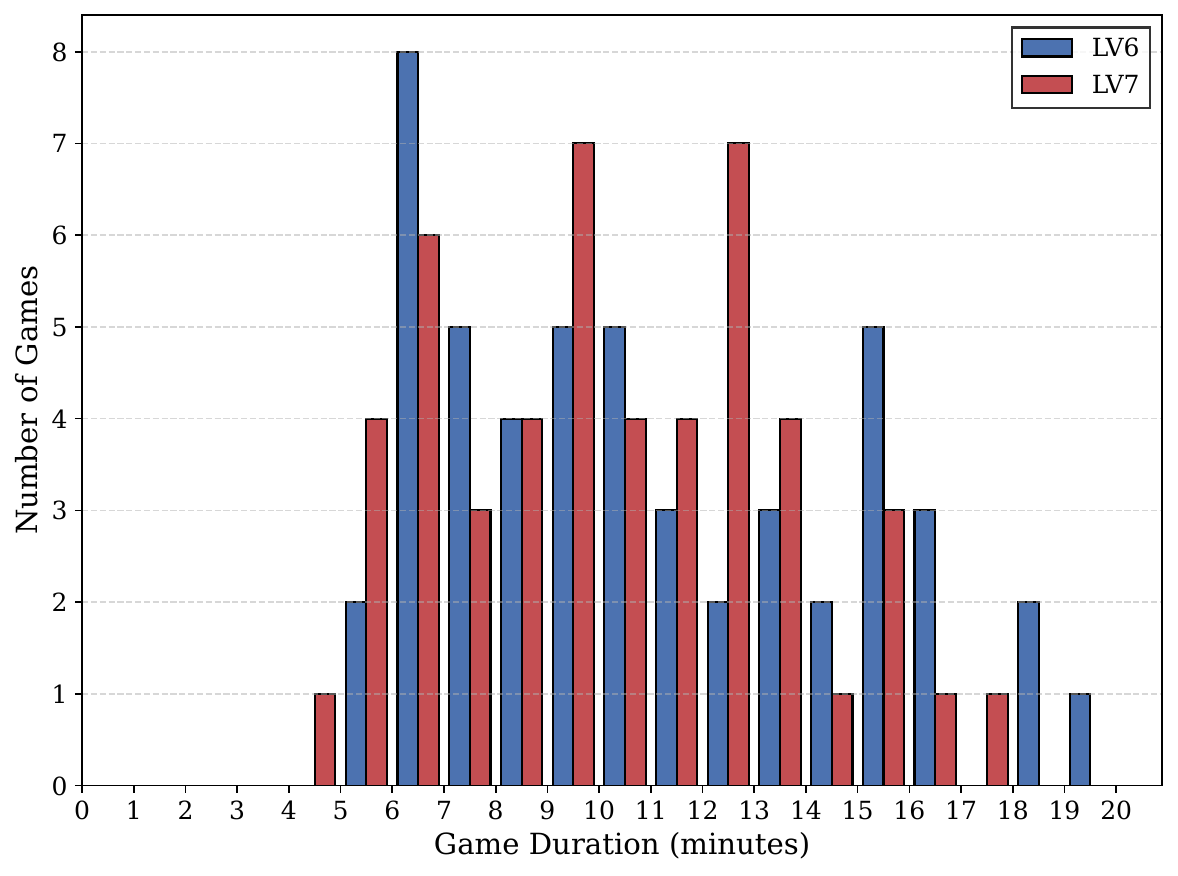} 
    \caption{Distribution of trajectory durations in the 100 collected games used to build \emph{SC2-Dynamics-50K}. For LV6, durations range from 347s to 1182s (mean 658.9s, median 606.0s). For LV7, durations range from 282s to 1033s (mean 616.3s, median 599.5s).}
    \label{fig:trajectory_duration_hist}
\end{figure}

\paragraph{Dataset Diversity and Coverage.}
Although \emph{SC2-Dynamics-50K} is collected in a controlled TvT-on-Flat64 setting, diversity is introduced through randomized strategy styles on both the scripted agent and the built-in AI, as well as variable game outcomes and game durations. 
\textbf{Across the 100 collected trajectories, the scripted agent achieves 56\% win rate against LV6 and 60\% against LV7}, indicating that the dataset contains both winning and losing games under different opponent difficulties. Figure~\ref{fig:trajectory_duration_hist} further shows that the collected trajectories span a broad range of game durations. While \emph{SC2-Dynamics-50K} does not cover the full SC2 state-action space, it spans a diverse and non-trivial subset of the TvT-on-Flat64 state-action space, which is sufficient for testing the core claims of this work.

\paragraph{Example of SC2-Dynamics-50K.}Appendix~\ref{appendix:Example of SC2-Dynamics-50K} provides a detailed sample of \emph{SC2-Dynamics-50K}.

\subsection{Offline Evaluation Settings.}\label{appendix:Offline Evaluation Settings.}
For the macro-situation metric (AWD), we set the penalty $\lambda=90.5$, which corresponds to the diagonal distance of the Flat64 map. For micro-entity matching, we set the spatial threshold $\delta=10.0$. We compare StarWM against three baselines: Static Bias (copying the input observation as the prediction), Qwen3-8B (zero-shot), and Qwen3-32B (zero-shot).

\subsection{Online Evaluation Settings.}\label{appendix:Online Testing Settings.}

\begin{table*}[t]
\centering
\small
\setlength{\tabcolsep}{4pt}
\renewcommand{\arraystretch}{1.28}
\rowcolors{2}{gray!10}{white}
\begin{tabular}{>{\centering\arraybackslash}m{2.0cm}
                >{\centering\arraybackslash}m{6.5cm}
                m{6.9cm}}
\toprule
\textbf{Name} & \textbf{Formula} & \textbf{Description} \\
\midrule

Win Rate
& $\displaystyle \mathrm{WR}=\dfrac{\text{Number of Games Won}}{\text{Total Number of Games Played}}\times 100\%$
& \textbf{Primary Metric}: Measures the proportion of games won, reflecting the overall performance of the agent. Higher values indicate stronger end-to-end performance. \\

Supply Block Rate
& $\displaystyle \mathrm{SBR}=\dfrac{\text{Time Supply Blocked}}{\text{Total Game Time}}\times 100\%$
& \textbf{Macro-Management Metric}: Measures the fraction of in-game time during which the agent is supply-blocked, reflecting its ability to balance production and supply expansion. Lower values indicate better macro management. \\

Resource Conversion Rate
& $\displaystyle \mathrm{RCR}=\dfrac{\text{Total Resources Spent}}{\text{Total Resources Collected}}\times 100\%$
& \textbf{Macro-Management Metric}: Measures the proportion of collected resources that are effectively spent, reflecting the agent's ability to translate economic growth into actions. Higher values indicate more efficient resource utilization. \\

Kill--Loss Ratio
& $\displaystyle \mathrm{KLR}=\dfrac{\text{Killed Enemy Army Value}}{\text{Lost Army Value}}\times 100\%$
& \textbf{Combat Metric}: Evaluates combat efficiency based on the economic value of killed and lost army units. Higher values indicate more favorable resource exchanges in combat. \\

Valid Action Rate
& $\displaystyle \mathrm{VAR}=\dfrac{\text{Number of Valid Actions}}{\text{Total Number of Issued Actions}}\times 100\%$
& \textbf{System Stability Metric}: Measures the proportion of valid actions issued by the agent, reflecting the robustness of action generation under diverse situations. An action is considered invalid if it cannot be executed by the game engine at the corresponding timestep. \\

Action Revision Rate
& $\displaystyle \mathrm{ARR}=\dfrac{1}{N}\sum_{t=1}^{N}\mathbb{I}\!\left(a_t^{\mathrm{final}}\neq a_t^{\mathrm{init}}\right)\times 100\%$
& \textbf{Mechanism Analysis Metric}: Measures how often the agent modifies its initial action proposal during decision refinement. Here, $a_t^{\mathrm{init}}$ and $a_t^{\mathrm{final}}$ denote the initial and final actions at decision step $t$, respectively, and $N$ is the total number of decision steps. This metric reflects the extent to which world model predictions influence final decisions. \\

\bottomrule
\end{tabular}
\caption{Detailed definitions and formulas of the online evaluation metrics.}
\label{tab:online-metrics}
\end{table*}

\paragraph{Implementation.}
We adopt SC2Arena~\cite{SC2Arena} as the online evaluation framework, with an LLM serving as the policy model. Online baselines include zero-shot Qwen3-8B/Qwen3-32B policies and a Chain-of-Summarization (CoS) baseline adapted from TextStarCraft II~\cite{textstarcraft2}. Since the full TextStarCraft II system is tied to a Protoss-vs-Zerg setting and relies on hand-crafted rule scripts for micro-control, it cannot be directly transferred to our setting. We therefore implemented a TextStarCraft-II-style CoS baseline in our platform using the same CoS prompt template as TextStarCraft II: it first summarizes the three most recent observations into a concise CoS summary, then generates executable actions conditioned on the current observation and this summary.

To implement our \emph{StarWM-Agent}, we extend SC2Arena with additional modules for world model prediction and action refinement, while reusing its original components for initial action generation (including instructions) and interaction with the underlying game engine. We conduct online matches against the SC2 built-in AI at Hard (LV5), Harder (LV6), and VeryHard (LV7), with the built-in AI's strategy style randomized, which represent the highest non-cheating difficulty levels. Due to resource constraints, all online experiments are conducted under the /no\_think setting.

\paragraph{Metrics for Online Evaluation.}We define multiple online metrics to characterize decision quality from multiple aspects, including overall performance, macro management, combat efficiency, system stability, and decision refinement. Table~\ref{tab:online-metrics} presents the detailed definitions and formulas of these metrics.

\section{Detailed Results of Offline Evaluation}

\subsection{Quantitative Results}\label{appendix:Detailed_Offline_Results}
Table~\ref{tab:detailed_offline_results} reports detailed quantitative results of offline evaluation, aligning with the conclusions presented in Table~\ref{tab:offline_results}.

\begin{table*}[htb]
\centering
\resizebox{\linewidth}{!}{
\begin{tabular}{llcccc}
\toprule
\textbf{Metric Classification} & \textbf{Metrics} & \textbf{Our StarWM} & \textbf{Qwen3-32B (Zero-shot)} & \textbf{Qwen3-8B (Zero-shot)} & \textbf{Static Bias} \\ \midrule
{\begin{tabular}[c]{@{}l@{}}Macro-Situation\\ Consistency\end{tabular}} 
 & Self Unit AWD & \textbf{5.96} & 13.44 & 14.41 & 13.89 \\
 & Self Struct AWD & \textbf{0.96} & 6.15 & 7.32 & 2.84 \\
 & Enemy Unit AWD & 30.94 & 27.75 & 28.34 & \textbf{27.23} \\
 & Enemy Struct AWD & 15.17 & 15.22 & 15.73 & \textbf{14.09} \\
 & Snapshot Enemy Struct AWD & 8.16 & 10.36 & 9.2 & \textbf{7.07} \\ \midrule
{\begin{tabular}[c]{@{}l@{}}Economy \&\\ Status\end{tabular}} 
 & Minerals SMAPE & \textbf{0.19} & 0.48 & 0.49 & 0.49 \\
 & Minerals Rate SMAPE & \textbf{0.09} & 0.73 & 0.37 & 0.12 \\
 & Gas SMAPE & \textbf{0.09} & 0.26 & 0.27 & 0.25 \\
 & Gas Rate SMAPE & \textbf{0.12} & 0.58 & 0.34 & \textbf{0.12} \\
 & Supply Used SMAPE & \textbf{0.02} & 0.21 & 0.04 & 0.04 \\
 & Supply Cap SMAPE & \textbf{0.01} & 0.2 & 0.03 & 0.02 \\
 & Alerts F1 & 0 & \textbf{0.02} & \textbf{0.02} & \textbf{0.02} \\
 & Upgrades F1 & \textbf{1} & 0.98 & 0.98 & 0.98 \\
 & Workers num SMAPE & \textbf{0.02} & 0.07 & 0.04 & 0.03 \\ \midrule
{Development} 
 & Queue F1 & \textbf{0.92} & 0.7 & 0.73 & 0.8 \\
 & Progress(\%) MAE & \textbf{0.43} & 24.27 & 27.07 & 24.81 \\ \midrule
{\begin{tabular}[c]{@{}l@{}}Micro Entity\\ Attributes\end{tabular}} 
 & Self Unit F1 & \textbf{0.94} & 0.87 & 0.85 & 0.85 \\
 & Self Unit Prec. & \textbf{0.94} & 0.87 & 0.86 & 0.86 \\
 & Self Unit Recall & \textbf{0.96} & 0.91 & 0.86 & 0.85 \\
 & Self Unit HP(\%) MAE & \textbf{4.15} & 5.11 & 5.62 & 4.81 \\
 & Self Unit Energy(\%) MAE & \textbf{1.47} & 4.81 & 3.89 & 3.49 \\
 & Self Struct F1 & \textbf{0.99} & 0.93 & 0.92 & 0.97 \\
 & Self Struct Prec. & \textbf{0.99} & 0.92 & 0.91 & 0.98 \\
 & Self Struct Recall & \textbf{0.99} & 0.96 & 0.94 & 0.96 \\
 & Self Struct HP(\%) MAE & 0.65 & 0.63 & 0.66 & \textbf{0.61} \\
 & Self Struct Energy(\%) MAE & \textbf{0.07} & 3.42 & 4.71 & 3.33 \\
 & Enemy Unit F1 & 0.65 & 0.69 & 0.69 & \textbf{0.7} \\
 & Enemy Unit Prec. & 0.68 & 0.73 & 0.72 & \textbf{0.74} \\
 & Enemy Unit Recall & 0.68 & 0.72 & 0.72 & \textbf{0.73} \\
 & Enemy Unit HP(\%) MAE & \textbf{7.9} & 8.47 & 9.01 & 9.15 \\
 & Enemy Unit Energy(\%) MAE & \textbf{2.54} & 2.59 & 2.61 & 2.58 \\
 & Enemy Struct F1 & 0.83 & 0.83 & 0.83 & \textbf{0.84} \\
 & Enemy Struct Prec. & 0.84 & 0.84 & 0.84 & \textbf{0.85} \\
 & Enemy Struct Recall & 0.84 & 0.84 & 0.83 & \textbf{0.85} \\
 & Enemy Struct HP(\%) MAE & \textbf{2.91} & 4.4 & 4.51 & 4.85 \\ 
 & Snap Enemy Struct F1 & 0.91 & 0.88 & 0.9 & \textbf{0.92} \\
 & Snap Enemy Struct Prec. & 0.92 & 0.9 & 0.91 & \textbf{0.93} \\
 & Snap Enemy Struct Recall & 0.91 & 0.89 & 0.9 & \textbf{0.93} \\ 
 \bottomrule
\end{tabular}
}
\caption{Detailed offline evaluation results across multiple dimensions.}
\label{tab:detailed_offline_results}
\end{table*}

\begin{table*}[htb]
    \centering
    \resizebox{\linewidth}{!}{
    \begin{tabular}{lcccc}
        \toprule
        Method & 
        \begin{tabular}[c]{@{}c@{}}Economy\\ (Minerals / Gas SMAPE$\downarrow$)\end{tabular} & 
        \begin{tabular}[c]{@{}c@{}}Development\\ (Queue F1$\uparrow$ / Progress MAE\%$\downarrow$)\end{tabular} & 
        \begin{tabular}[c]{@{}c@{}}Micro-Entity\\ (Self / Enemy HP MAE\%$\downarrow$)\end{tabular} & 
        \begin{tabular}[c]{@{}c@{}}Macro-Situation\\ (Self / Enemy AWD$\downarrow$)\end{tabular} \\
        \midrule
        StarWM         & 0.19 / 0.09 & 0.92 / 0.43 & 4.15 / 7.90 & 3.46 / 18.09 \\
        StarWM-History & 0.19 / \textbf{0.08} & 0.92 / \textbf{0.21} & \textbf{4.08 / 7.84} & \textbf{3.21 / 17.87} \\
        \bottomrule
    \end{tabular}
    }
    \caption{Offline evaluation of single-frame StarWM and history-augmented StarWM. The history-augmented variant uses three recent observations $(o_{t-2}, o_{t-1}, o_t)$ together with the action sequence $a_{t:t+\tau}$ to predict $o_{t+\tau}$.}
    \label{tab:history_augmented_starwm}
\end{table*}

\subsection{History-Augmented StarWM under Partial Observability}
\label{appendix:history_augmented_starwm}

To explore whether temporal information helps world-model prediction under partial observability, we further train a history-augmented StarWM with $h=3$, using $(o_{t-2}, o_{t-1}, o_t, a_{t:t+\tau})$ to predict $o_{t+\tau}$. Table~\ref{tab:history_augmented_starwm} compares the single-frame StarWM with its history-augmented variant.

As shown in Table~\ref{tab:history_augmented_starwm}, temporal history improves overall prediction performance. In particular, Progress MAE decreases from 0.43\% to 0.21\%, and self-side AWD decreases from 3.46 to 3.21, indicating better task-progress estimation and self-side macro-situation consistency. Enemy-side AWD also improves slightly from 18.09 to 17.87, but remains challenging under Fog of War. These results suggest that temporal information is useful for world modeling, while accurate prediction under partial observability likely requires stronger memory mechanisms for longer-history compression and belief-state modeling, which we leave for future work.

\subsection{Temporal and Qualitative Analyses}\label{appendix:Temporal_and_Qualitative Analyses}

\begin{figure*}[htb]
    \centering
    \includegraphics[width=\linewidth]{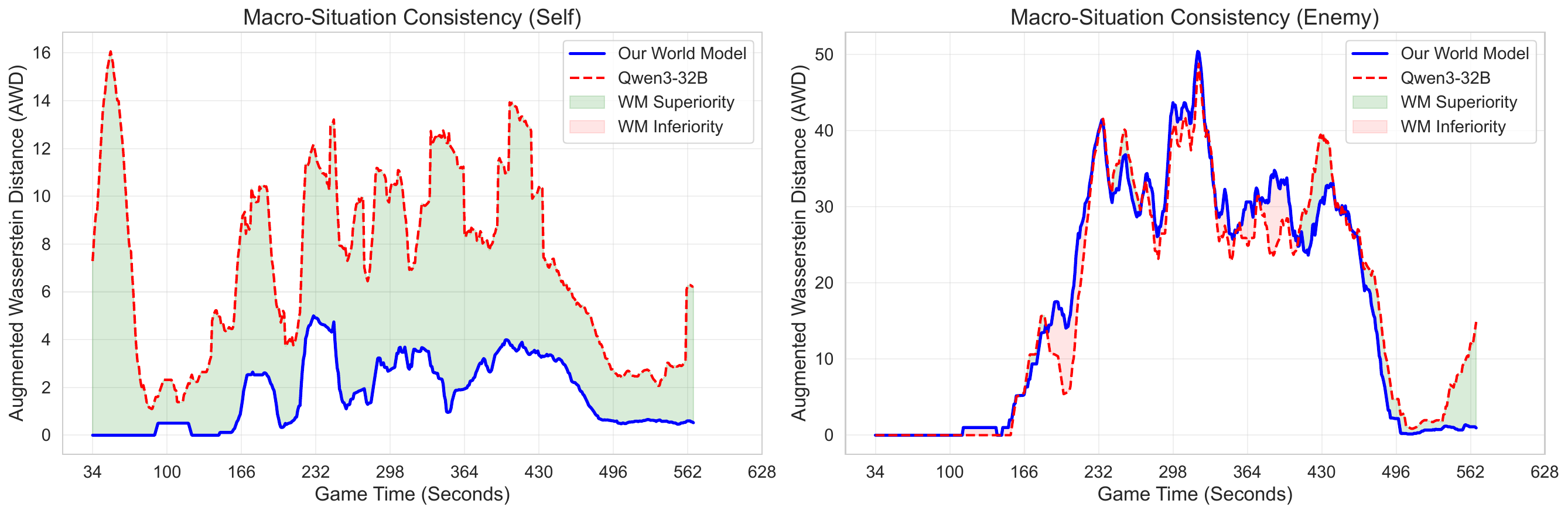} 
    \caption{Evolution of Macro-Situation Metric (AWD) over game time. Left: Self-side entities. Right: Enemy-side entities. The green area indicates where StarWM outperforms the zero-shot Qwen3-32B baseline.}
    \label{fig:dynamics_split}
\end{figure*}

\paragraph{Temporal Analysis of Macro-Situation Consistency.}
Figure~\ref{fig:dynamics_split} shows the temporal evolution of AWD over game time. StarWM maintains consistently lower self-side AWD than the zero-shot 32B baseline, indicating more stable macro-situation prediction.

\begin{figure*}[htb]
    \centering
    \includegraphics[width=\linewidth]{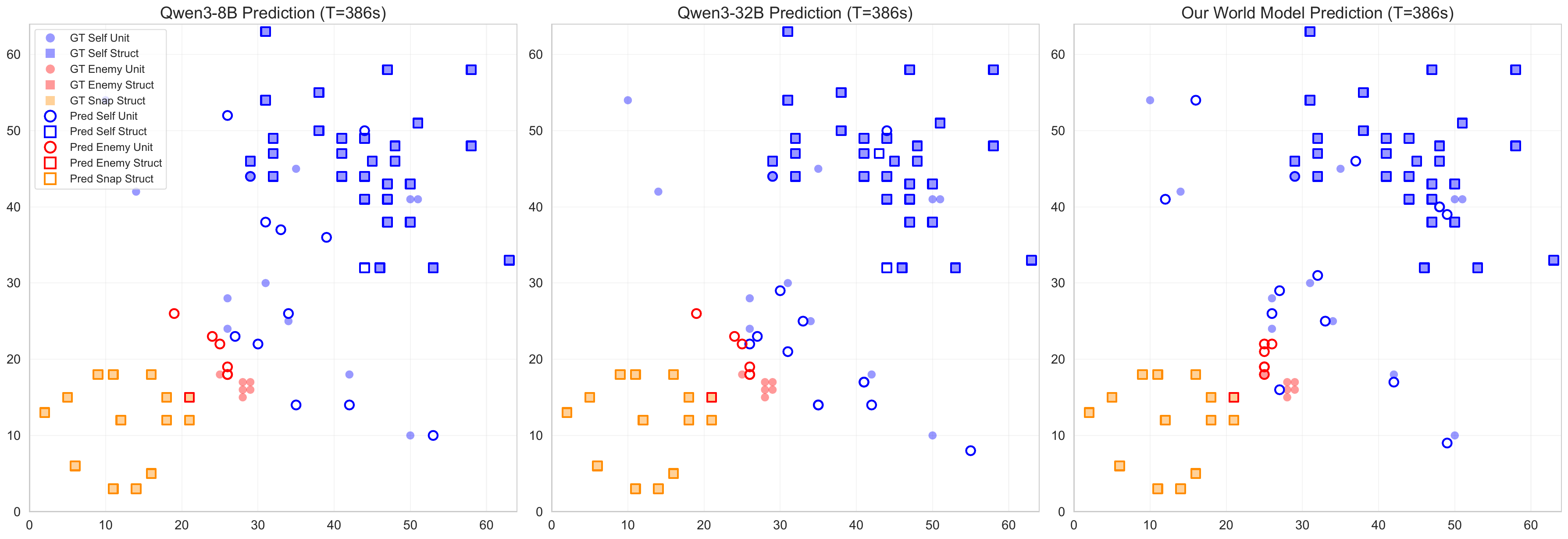}
    \caption{Offline case study. Left: Qwen3-8B. Middle: Qwen3-32B. Right: StarWM. Circles and squares denote units and structures, respectively. Filled markers indicate ground truth, while hollow markers indicate predictions. The spatial distribution of self-side units predicted by StarWM (hollow circles) closely matches the ground truth (filled circles) and preserves the army formation, whereas zero-shot LLM predictions are more scattered and less coherent.}
    \label{fig:case_study}
\end{figure*}

\begin{figure*}[htb]
    \centering
    \includegraphics[width=\linewidth]{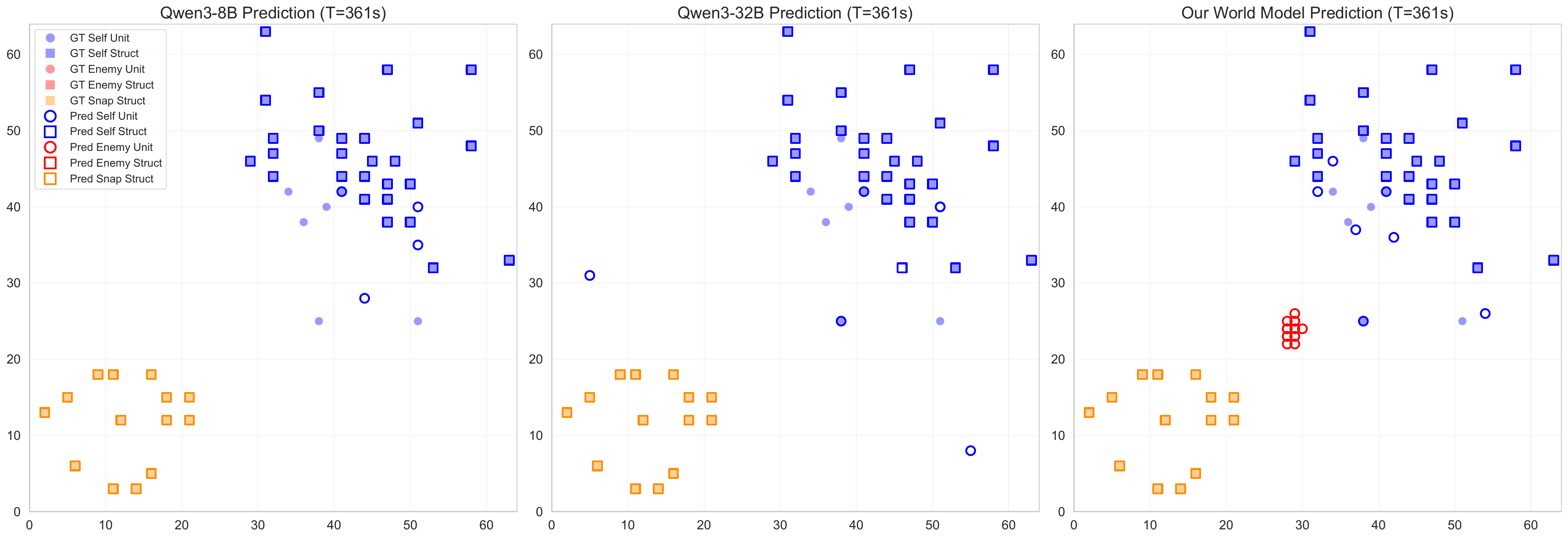}
    \caption{Offline case study under Fog of War. When self-side units enter unobservable regions, StarWM predicts potential enemy presence within those areas (red hollow circles). This illustrates a plausible statistical predictive pattern under partial observability.}
    \label{fig:fog_hallucination}
\end{figure*}

\paragraph{Case Study.}Figure~\ref{fig:case_study} provides an intuitive example. The spatial distribution of self units predicted by StarWM (hollow circles) closely matches the ground truth (filled circles), preserving the army formation, whereas zero-shot LLM predictions appear more scattered and less coherent. This is consistent with StarWM's lower self-side AWD in Table~\ref{tab:offline_results}.
Notably, Figure~\ref{fig:fog_hallucination} presents an interesting phenomenon. When self units move into unobservable areas, StarWM predicts potential enemy units within those regions (red hollow circles). Although this causes a false positive in offline evaluation, it suggests the learned statistical regularity that enemy defenders are likely to exist when approaching hostile territory.

\subsection{Cross-Map and Out-of-Domain Opponent Probe}\label{appendix:Cross-Map and Out-of-Domain Opponent Probe}
To assess whether the dynamics learned by StarWM transfer beyond the training map and opponent distribution, we evaluate StarWM on a small out-of-domain test set without retraining. Specifically, we collect 5 trajectories on the unseen Flat128 map, using the same rule-based agent against the SC2 built-in AI at Hard (LV5), with randomized strategy styles on both sides. Using the same sample construction procedure as in SC2-Dynamics-50K, with a 5-second prediction horizon and a 1-second sliding step, we obtain 2,969 test samples.

Despite the map and opponent distribution shift, the overall trend in Table~\ref{tab:cross_map_probe} remains broadly consistent with Table~\ref{tab:offline_results}: StarWM still performs best on most economy, development, and self-side metrics, while Static Bias remains strongest on Enemy AWD. This suggests that StarWM learns dynamics that are not merely tied to the original training distribution, but generalize to some extent across map and opponent shifts.

\begin{table*}[tb]
    \centering
    \resizebox{\linewidth}{!}{
    \begin{tabular}{lcccc}
        \toprule
        Method & 
        \begin{tabular}[c]{@{}c@{}}Economy\\ (Minerals / Gas SMAPE$\downarrow$)\end{tabular} & 
        \begin{tabular}[c]{@{}c@{}}Development\\ (Queue F1$\uparrow$ / Progress MAE\%$\downarrow$)\end{tabular} & 
        \begin{tabular}[c]{@{}c@{}}Micro-Entity\\ (Self / Enemy HP MAE\%$\downarrow$)\end{tabular} & 
        \begin{tabular}[c]{@{}c@{}}Macro-Situation\\ (Self / Enemy AWD$\downarrow$)\end{tabular} \\
        \midrule
        Static Bias & 0.66 / 0.43 & 0.85 / 26.16 & 2.07 / 14.48 & 5.29 / \textbf{18.23} \\
        Qwen3-8B    & 0.65 / 0.40 & 0.79 / 28.27 & 2.34 / 14.40 & 6.59 / 18.76 \\
        Qwen3-32B   & 0.67 / 0.43 & 0.75 / 25.16 & 2.28 / 13.44 & 7.01 / 19.57 \\
        \textbf{StarWM} & \textbf{0.28 / 0.14} & \textbf{0.95 / 0.35} & \textbf{1.85 / 11.88} & \textbf{2.55} / 22.02 \\
        \bottomrule
    \end{tabular}
    }
    \caption{Cross-map and out-of-domain opponent probe without retraining. All methods are tested on 2,969 samples constructed from 5 trajectories on the unseen Flat128 map against the SC2 built-in AI at Hard (LV5). StarWM remains strongest on most economy, development, and self-side metrics, while Enemy AWD remains challenging.}
    \label{tab:cross_map_probe}
\end{table*}

\subsection{Action-Conditioned Prediction Probe}\label{appendix:Action-Conditioned_Prediction_Probe}
\begin{figure*}[t]
    \centering
    \setlength{\tabcolsep}{4pt}
    \renewcommand{\arraystretch}{1.0}
    \begin{tabular}{cc}
        \textbf{Current Observation} & \textbf{Future Observation (No Action)} \\[-1mm]
        \includegraphics[width=0.37\linewidth,height=0.28\textheight]{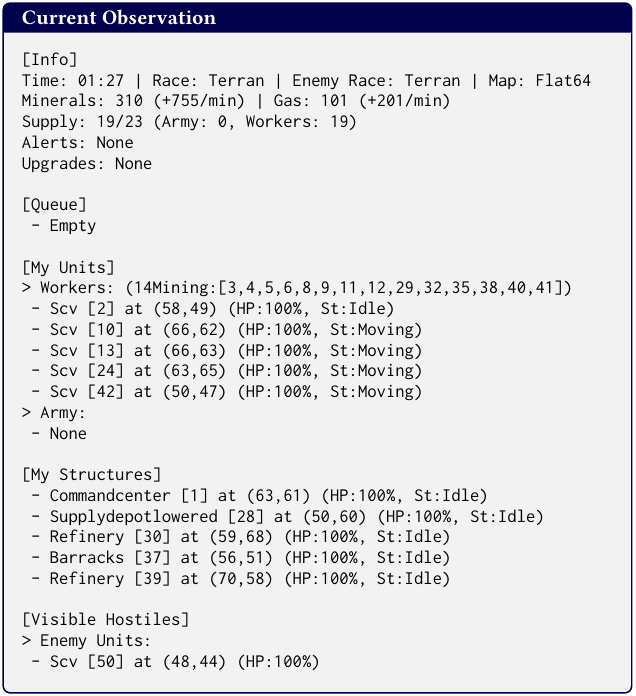} &
        \includegraphics[width=0.37\linewidth,height=0.28\textheight]{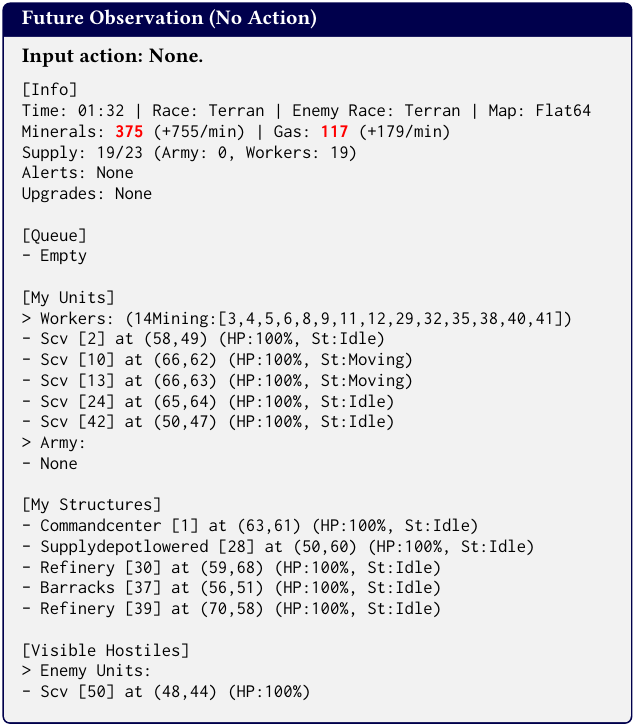} \\[1mm]

        \textbf{Future Observation (Train SCV)} & \textbf{Future Observation (Train Marine)} \\[-1mm]
        \includegraphics[width=0.37\linewidth,height=0.28\textheight]{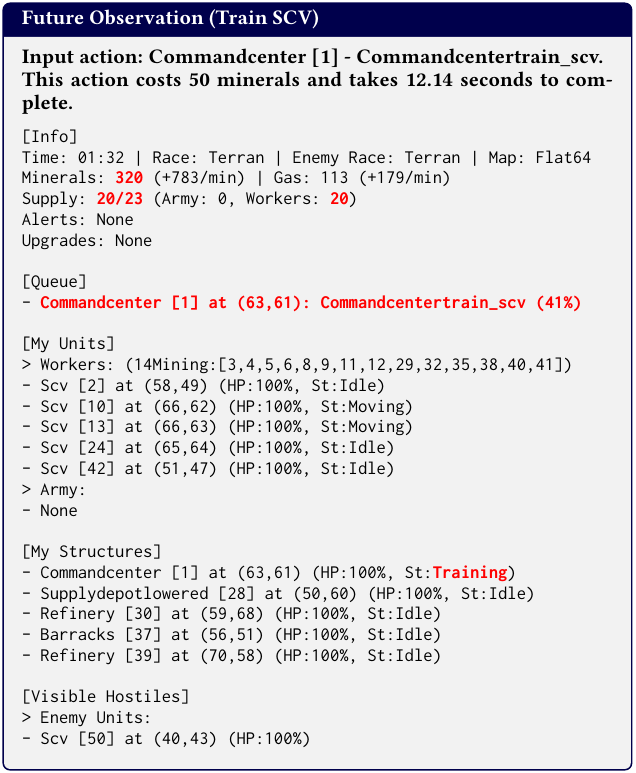} &
        \includegraphics[width=0.37\linewidth,height=0.28\textheight]{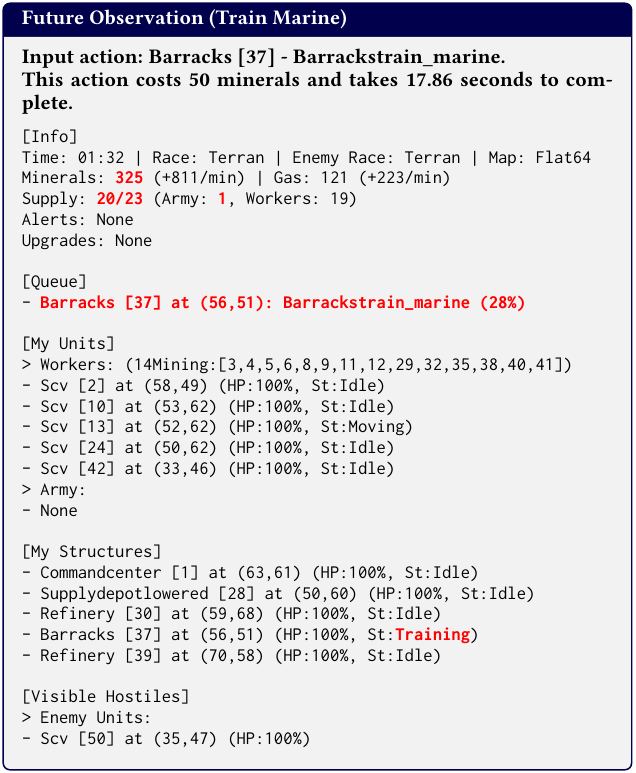} \\[1mm]

        \textbf{Future Observation (Build Factory)} & \textbf{Future Observation (Attack)} \\[-1mm]
        \includegraphics[width=0.37\linewidth,height=0.28\textheight]{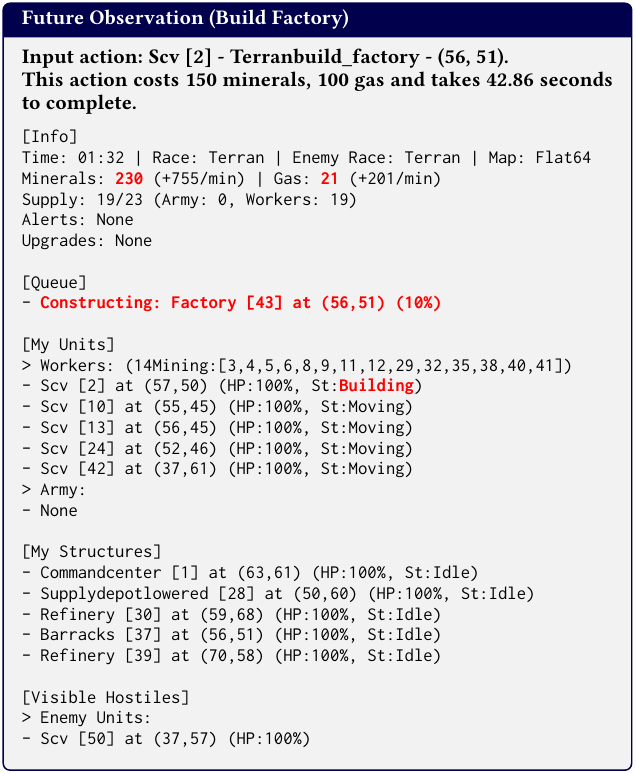} &
        \includegraphics[width=0.37\linewidth,height=0.28\textheight]{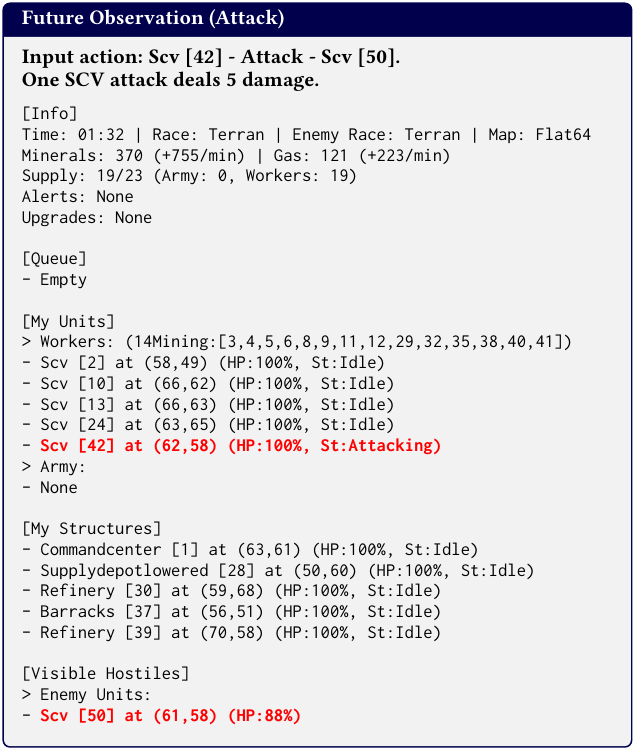} \\
    \end{tabular}
    \caption{Action-conditioned prediction probe of StarWM. The top-left figure shows the shared current observation, and the other five figures show predicted future observations 5 seconds later under different single-action inputs. Action-relevant changes are highlighted in \textcolor{red}{red}. Different actions lead to distinct predicted future observations.}
    \label{fig:action_conditioned_case_study}
\end{figure*}

We present a case study in which we fix an early-game observation and compare the predicted future observations under five different single-action inputs. Figure~\ref{fig:action_conditioned_case_study} shows the corresponding future observations predicted by StarWM 5 seconds later, with action-relevant changes highlighted in red.

The examples show that different actions lead to distinct and logically consistent future predictions. Production and construction actions mainly induce corresponding changes in resources, queue, and structure status, while combat actions mainly affect unit status and health. This provides a direct qualitative illustration that StarWM learns action-conditioned dynamics, rather than merely reproducing surface-level textual patterns.

\subsection{Prediction Horizon Analysis.}\label{appendix:Prediction Horizon Analysis.}
We choose a 5-second prediction horizon for our main experiments because even a 5-second interval in SC2 can contain decision-critical changes: resources may be consumed or become insufficient, supply may approach its cap, production tasks may complete, buildings may become idle, units may move to tactically different positions, and combat may change unit health, survival, and action feasibility.

To explore a longer prediction horizon, we further train StarWM with $\tau=10$ under the same training-data scale as the $\tau=5$ setting.

\begin{table*}[tb]
    \centering
    \resizebox{\linewidth}{!}{
    \begin{tabular}{lcccc}
        \toprule
        Method & 
        \begin{tabular}[c]{@{}c@{}}Economy\\ (Minerals / Gas SMAPE$\downarrow$)\end{tabular} & 
        \begin{tabular}[c]{@{}c@{}}Development\\ (Queue F1$\uparrow$ / Progress MAE\%$\downarrow$)\end{tabular} & 
        \begin{tabular}[c]{@{}c@{}}Micro-Entity\\ (Self / Enemy HP MAE\%$\downarrow$)\end{tabular} & 
        \begin{tabular}[c]{@{}c@{}}Macro-Situation\\ (Self / Enemy AWD$\downarrow$)\end{tabular} \\
        \midrule
        Static Bias, $\tau=5$ & 0.49 / 0.25 & 0.80 / 24.81 & 4.81 / 9.15 & 8.37 / \textbf{16.13} \\
        \textbf{StarWM}, $\tau=5$ & \textbf{0.19 / 0.09} & \textbf{0.92 / 0.43} & \textbf{4.15 / 7.90} & \textbf{3.46} / 18.09 \\
        \midrule
        Static Bias, $\tau=10$ & 0.58 / 0.32 & 0.68 / 31.80 & 6.76 / \textbf{12.46} & 12.90 / \textbf{21.63} \\
        \textbf{StarWM}, $\tau=10$ & \textbf{0.28 / 0.14} & \textbf{0.88 / 0.69} & \textbf{5.36} / 12.60 & \textbf{5.48} / 23.12 \\
        \bottomrule
    \end{tabular}
    }
    \caption{Prediction-horizon analysis. $\tau$ denotes the prediction horizon. Bold indicates the better result within each horizon.}
    \label{tab:prediction_horizon}
\end{table*}

As shown in Table~\ref{tab:prediction_horizon}, Static Bias degrades substantially from $\tau=5$ to $\tau=10$, indicating that larger state changes accumulate over the longer interval. StarWM at $\tau=10$ also degrades relative to $\tau=5$, while still substantially outperforming Static Bias on most economy, development, and self-side metrics.

These results suggest that the current formulation provides effective short-horizon prediction, while longer-horizon prediction does not completely collapse but becomes considerably more difficult. More accurate longer-horizon prediction may require multi-scale hierarchical prediction or temporal abstraction, which represent important directions for future work.

\clearpage
\raggedbottom

\section{Detailed Results of Online Evaluation}

\subsection{Robustness of Online Win Rates.}
Due to the high computational cost of online evaluation, our main experiments use 20 matches for each setting. To further assess the robustness of the observed improvement, we additionally expand the direct comparison between Qwen3-8B and StarWM-Agent (8B) on LV7 (VeryHard) to 100 matches per method under the same protocol, including the original 20 matches.

\begin{table*}[htb]
    \centering
    \resizebox{\linewidth}{!}{
    \begin{tabular}{lcccccc}
        \toprule
        Setting &
        Wins/Games &
        Win Rate (95\% Wilson CI)$\uparrow$ &
        SBR (\%)$\downarrow$ &
        RCR (\%)$\uparrow$ &
        KLR (\%)$\uparrow$ &
        VAR (\%)$\uparrow$ \\
        \midrule
        Qwen3-8B &
        5/100 &
        5.0\% [2.2\%, 11.2\%] &
        39.65 $\pm$ 25.10 &
        44.56 $\pm$ 30.06 &
        13.85 &
        48.73 $\pm$ 27.27 \\
        
        \textbf{StarWM-Agent (8B)} &
        23/100 &
        \textbf{23.0\% [15.8\%, 32.2\%]} &
        \textbf{6.08 $\pm$ 7.24} &
        \textbf{82.47 $\pm$ 9.57} &
        \textbf{22.09} &
        \textbf{84.35 $\pm$ 12.07} \\
        \bottomrule
    \end{tabular}
    }
    \caption{Expanded online evaluation against the SC2 built-in AI at LV7 (VeryHard). Each method is evaluated over 100 matches.}
    \label{tab:expanded_online_evaluation}
\end{table*}

As shown in Table~\ref{tab:expanded_online_evaluation}, StarWM-Agent (8B) achieves a 23.0\% win rate compared with 5.0\% for Qwen3-8B. Their 95\% Wilson confidence intervals are [15.8\%, 32.2\%] and [2.2\%, 11.2\%], respectively, and a two-sided Fisher exact test shows a statistically significant difference ($p=0.00037$). The improvements are also consistent across SBR, RCR, KLR, and VAR.

\subsection{Analysis of Action Revisions.}
\begin{figure*}[htb]
    \centering
    \includegraphics[width=\linewidth]{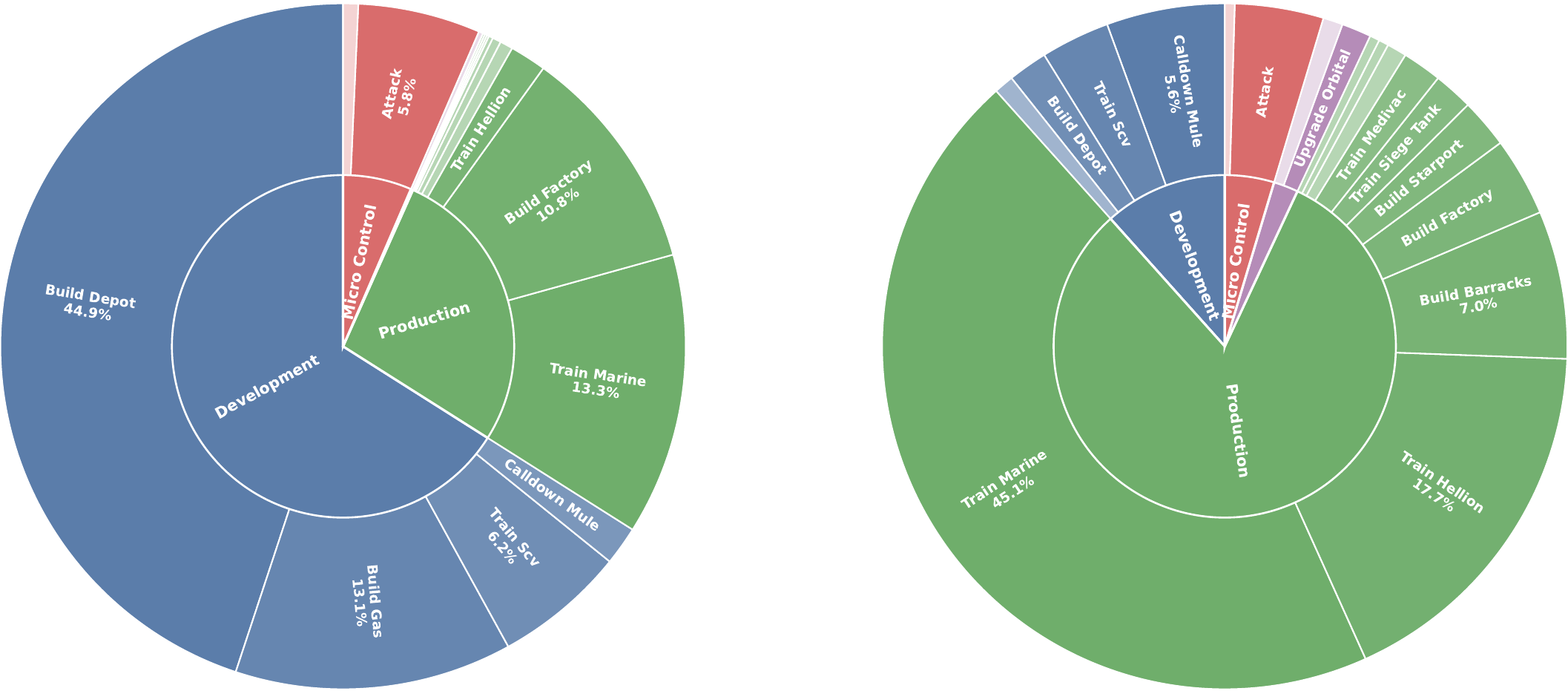}
    \caption{Distribution of action revision types for StarWM-Agent (32B) during online evaluation. Left: Added actions. Right: Removed actions.}
    \label{fig:action_analysis_32b}
\end{figure*}

Under the main settings (Table~\ref{tab:online_metrics}), the Action Revision Rate of StarWM-Agent (8B) and StarWM-Agent (32B) is 32.74\% and 19.45\%, respectively, aggregated across the three opponents. This indicates that world model simulations frequently lead to action revisions, especially for the smaller policy model. Figure~\ref{fig:action_analysis_32b} further breaks down the revision behavior of StarWM-Agent (32B) by action type. Among added actions, \emph{Build Supply Depot} accounts for the largest proportion (44.9\%), suggesting that world model simulations effectively promote stronger macro management.

\clearpage

\section{Examples of SC2-Dynamics-50K and StarWM-Agent Online Decision Making}

\subsection{Example of SC2-Dynamics-50K}\label{appendix:Example of SC2-Dynamics-50K}
We show a representative sample from the \emph{SC2-Dynamics-50K} dataset, including the current textual observation, the action sequence, and the target future observation.

\begin{tcolorbox}[
    colback=black!5!white,
    colframe=blue!30!black,
    title=\textbf{A sample from SC2-Dynamics-50K},
    fonttitle=\bfseries,
    arc=1mm,
    boxrule=0.8pt,
    width=\linewidth,
    left=2mm,
    right=2mm,
    top=1mm,
    bottom=0mm,
    breakable
]
\textbf{Input:}
\begin{lstlisting}[style=promptstyle]
{
    "role": "user",
    "content": "### Role
You are the StarCraft II World Model, a dynamics simulator for Player 1.

### Task
Predict the Future Observation for Player 1 exactly 5 seconds later based on the Current Observation, the Action Sequence and the Game Mechanics.

### Input
1. Current Observation: The partial state visible to Player 1 at time t.
2. Action Sequence: All actions executed by Player 1 between t and t+5s.

### Output
Future Observation: The Predicted partial state visible to Player 1 at time t+5s.

### Data Explanation
#### Observation Schema:
- **[Info]**: Player-specific status, race, enemy race, economy (minerals, gas, gather rate), supply (used, cap), alerts and upgrades.
- **[Queue]**: Ongoing and queued production, construction, and research tasks.
- **[My Units]**: Self units (workers & army).
- **[My Structures]**: Self buildings, including key structures for production and technology.
- **[Visible Hostiles]**: Enemy units and structures currently visible within the Fog of War.
#### Action Schema:
The Action follows the format: Subject - Command - Target.
- **Subject**: UnitName [ID] (e.g., Marine [101]).
- **Command**: The specific operation (e.g., Move, Attack, etc.).
- **Target**: Coordinate (x, y) or TargetName [ID] or absent for some commands.

### Input Data
#### Current Observation:
[Info]
Time: 05:44 | Race: Terran | Enemy Race: Terran | Map: Flat64
Minerals: 280 (+2575/min) | Gas: 124 (+223/min)
Supply: 64/93 (Army: 16, Workers: 48)
Alerts: None
Upgrades: Punishergrenades

[Queue]
 - Orbitalcommand [1] at (63,61): Commandcentertrain_scv (34%)
 - Barracks [65] at (53,59): Barrackstrain_marine (80%)
 - Orbitalcommand [83] at (43,64): Commandcentertrain_scv (93%)
 - Barracks [86] at (55,55): Barrackstrain_marine (25%)
 - Barracks [90] at (58,57): Barrackstrain_marine (53%), Barrackstrain_marine (35%)
 - Commandcenter [226] at (24,61): Upgradetoorbital_orbitalcommand (0%)
 - Constructing: Barracksreactor [334] at (65,54) (59%)
 - Constructing: Refinery [335] at (28,68) (94%)
 - Constructing: Missileturret [352] at (18,68) (65%)

[My Units]
> Workers: (38Mining:[5,6,7,8,9,10,11,13,35,61,71,77,84,115,139,
144,153,155,159,160,179,183,196,217,222,225,
237,240,254,256,271,
292,307,311,320,337,351,360], 2Mule:[305,313])
 - Scv [2] at (53,72) (HP:100%, St:Idle)
 - Scv [3] at (23,64) (HP:100%, St:Moving)
 - Scv [82] at (45,59) (HP:100%, St:Building)
 - Scv [154] at (21,65) (HP:100%, St:Moving)
 - Scv [264] at (28,69) (HP:100%, St:Building)
 - Scv [326] at (18,67) (HP:100%, St:Building)
> Army:
 - Marine [135] at (37,57) (HP:100%, St:Idle)
 - Marine [143] at (37,56) (HP:100%, St:Idle)
 - Marine [147] at (36,57) (HP:100%, St:Attacking)
 - Marine [310] at (40,27) (HP:100%, St:Attacking)
 - Marine [324] at (41,19) (HP:100%, St:Attacking)
 - Marauder [332] at (44,38) (HP:73%, St:Idle)
 - Marauder [347] at (33,39) (HP:100%, St:Attacking)
 - Marine [356] at (61,46) (HP:100%, St:Attacking)
 - Marine [358] at (46,53) (HP:100%, St:Attacking)
 - Marine [359] at (41,49) (HP:100%, St:Attacking)

[My Structures]
 - Orbitalcommand [1] at (63,61) (HP:100%, Eng:14%, St:Training)
 - Supplydepotlowered [36] at (50,60) (HP:100%, St:Idle)
 - Barracks [65] at (53,59) (HP:100%, St:Training, AddOn:Techlab [148])
 - Orbitalcommand [83] at (43,64) (HP:100%, Eng:17%, St:Training)
 - Barracks [86] at (55,55) (HP:100%, St:Training, AddOn:Techlab [150])
 - Barracks [90] at (58,57) (HP:100%, St:Training, AddOn:Reactor [158])
 - Refinery [114] at (70,58) (HP:100%, St:Idle)
 - Refinery [122] at (59,68) (HP:100%, St:Idle)
 - Bunker [140] at (44,54) (HP:100%, St:Idle)
 - Engineeringbay [146] at (57,52) (HP:100%, St:Idle)
 - Barrackstechlab [148] at (56,59) (HP:100%, St:Idle)
 - Barrackstechlab [150] at (58,55) (HP:100%, St:Idle)
 - Supplydepotlowered [156] at (53,57) (HP:100%, St:Idle)
 - Barracksreactor [158] at (61,57) (HP:100%, St:Idle)
 - Supplydepotlowered [164] at (60,53) (HP:100%, St:Idle)
 - Factory [190] at (43,57) (HP:100%, St:Idle, AddOn:Techlab [252])
 - Commandcenter [226] at (24,61) (HP:100%, St:Upgrading)
 - Supplydepotlowered [251] at (55,53) (HP:100%, St:Idle)
 - Factorytechlab [252] at (46,57) (HP:100%, St:Idle)
 - Barracks [257] at (62,54) (HP:100%, St:Building)
 - Supplydepotlowered [265] at (42,55) (HP:100%, St:Idle)
 - Missileturret [299] at (43,73) (HP:100%, St:Idle)
 - Missileturret [314] at (70,68) (HP:100%, St:Idle)
 - Supplydepot [333] at (46,59) (HP:100%, St:Idle)

[Visible Hostiles]
> Enemy Units:
 - Scv [88] at (29,30) (HP:100%)
 - Banshee [230] at (28,30) (HP:87%, Eng:3%)
 - Banshee [239] at (30,29) (HP:87%, Eng:15%)
> Enemy Structures:
 - Starporttechlab [162] at (33,25) (HP:92%)
 - Barrackstechlab [365] at (33,22) (HP:100%)
> Snapshot Enemy Structures:
 - Factory at (23,28)
 - Supplydepotlowered at (21,28)
 - Orbitalcommand at (24,22)
 - Refinery at (17,25)
 - Supplydepotlowered at (28,28)
 - Barracks at (30,22)
 - Refinery at (28,15)
 - Barracks at (30,28)
 - Starport at (30,25)


#### Action Sequence(5s):
- +0.1s: Marauder [347] - Attack - (31, 34)
- +0.2s: Scv [196] - Move - (45, 66)
- +0.2s: Scv [196] - Smartcmd - Orbitalcommand [83]
- +0.3s: Supplydepot [333] - Morph_supplydepot_lower
- +0.3s: Marauder [347] - Attack - (30, 34)
- +0.4s: Scv [240] - Move - (68, 64)
- +0.4s: Scv [240] - Smartcmd - Mineralfield [19]
- +0.4s: Scv [217] - Move - (65, 63)
- +0.4s: Scv [217] - Smartcmd - Orbitalcommand [1]
- +0.4s: Scv [144] - Move - (66, 62)
- +0.4s: Scv [144] - Smartcmd - Orbitalcommand [1]
- +0.4s: Scv [7] - Move - (64, 64)
- +0.4s: Scv [7] - Smartcmd - Orbitalcommand [1]
- +0.4s: Mule [313] - Move - (39, 68)
- +0.4s: Mule [313] - Smartcmd - Mineralfield750 [32]
- +0.4s: Marauder [347] - Attack - (30, 33)
- +0.6s: Factory [190] - Factorytrain_siegetank
- +0.6s: Scv [337] - Move - (19, 64)
- +0.6s: Scv [337] - Smartcmd - Mineralfield750 [46]
- +0.6s: Marauder [347] - Attack - (30, 33)
- +0.8s: Scv [237] - Move - (41, 66)
- +0.8s: Scv [237] - Smartcmd - Orbitalcommand [83]
- +0.8s: Marauder [347] - Attack - (30, 33)
- +0.9s: Orbitalcommand [83] - Commandcentertrain_scv
- +1.0s: Scv [367] - Harvest_gather - Mineralfield [33]
- +1.0s: Marauder [347] - Attack - (30, 32)
- +1.0s: Marine [135] - Attack - (36, 57)
- +1.0s: Marine [143] - Attack - (36, 57)
- +1.0s: Marine [147] - Attack - (36, 57)
- +1.2s: Marauder [347] - Move - (30, 33)
- +1.2s: Scv [13] - Move - (65, 63)
- +1.2s: Scv [13] - Smartcmd - Orbitalcommand [1]
- +1.2s: Scv [3] - Move - (22, 67)
- +1.2s: Scv [3] - Smartcmd - Mineralfield750 [42]
- +1.3s: Scv [196] - Move - (48, 67)
- +1.3s: Scv [196] - Smartcmd - Mineralfield [25]
- +1.3s: Marauder [347] - Move - (30, 33)
- +1.5s: Marauder [347] - Move - (30, 33)
- +1.6s: Scv [7] - Harvest_gather - Refinery [335]
- +1.6s: Scv [61] - Harvest_gather - Refinery [335]
- +1.6s: Scv [264] - Harvest_gather - Mineralfield750 [41]
- +1.6s: Scv [217] - Move - (66, 66)
- +1.6s: Scv [217] - Smartcmd - Mineralfield750 [21]
- +1.7s: Scv [144] - Move - (69, 63)
- +1.7s: Scv [144] - Smartcmd - Mineralfield750 [15]
- +1.7s: Scv [5] - Move - (69, 62)
- +1.7s: Scv [5] - Smartcmd - Mineralfield750 [16]
- +1.7s: Marauder [347] - Move - (32, 35)
- +1.9s: Marauder [347] - Move - (32, 35)
- +2.1s: Scv [237] - Move - (38, 67)
- +2.1s: Scv [237] - Smartcmd - Mineralfield [33]
- +2.1s: Marauder [347] - Move - (32, 35)
- +2.1s: Marine [135] - Attack - (36, 57)
- +2.1s: Marine [143] - Attack - (36, 57)
- +2.1s: Marine [147] - Attack - (36, 57)
- +2.2s: Marauder [347] - Attack - (26, 33)
- +2.3s: Scv [9] - Move - (45, 66)
- +2.3s: Scv [9] - Smartcmd - Orbitalcommand [83]
- +2.3s: Scv [6] - Move - (21, 61)
- +2.3s: Scv [6] - Smartcmd - Commandcenter [226]
- +2.4s: Scv [13] - Move - (68, 64)
- +2.4s: Scv [13] - Smartcmd - Mineralfield [19]
- +2.4s: Marauder [347] - Move - (30, 32)
- +2.4s: Marine [356] - Attack - (67, 18)
- +2.6s: Marauder [347] - Move - (32, 35)
- +2.7s: Scv [351] - Move - (21, 61)
- +2.7s: Scv [351] - Smartcmd - Commandcenter [226]
- +2.8s: Marauder [347] - Move - (33, 35)
- +2.9s: Mule [305] - Move - (42, 67)
- +2.9s: Mule [305] - Smartcmd - Orbitalcommand [83]
- +2.9s: Marauder [332] - Attack - (20, 22)
- +2.9s: Marauder [347] - Move - (33, 35)
- +3.0s: Marauder [332] - Move - (42, 37)
- +3.1s: Scv [311] - Move - (65, 63)
- +3.1s: Scv [311] - Smartcmd - Orbitalcommand [1]
- +3.1s: Marauder [347] - Move - (33, 35)
- +3.1s: Marine [135] - Attack - (36, 57)
- +3.1s: Marine [143] - Attack - (36, 57)
- +3.1s: Marine [147] - Attack - (36, 57)
- +3.2s: Scv [367] - Move - (38, 67)
- +3.2s: Scv [367] - Smartcmd - Mineralfield [33]
- +3.2s: Marauder [332] - Attack - (20, 22)
- +3.3s: Marauder [347] - Move - (33, 35)
- +3.3s: Scv [2] - Move - (50, 70)
- +3.3s: Scv [2] - Move - (53, 72)
- +3.3s: Marine [359] - Attack - (17, 41)
- +3.4s: Scv [10] - Move - (64, 64)
- +3.4s: Scv [10] - Smartcmd - Orbitalcommand [1]
- +3.4s: Marauder [332] - Move - (41, 37)
- +3.5s: Scv [9] - Move - (48, 67)
- +3.5s: Scv [9] - Smartcmd - Mineralfield [25]
- +3.5s: Scv [254] - Move - (24, 64)
- +3.5s: Scv [254] - Smartcmd - Commandcenter [226]
- +3.5s: Scv [6] - Move - (18, 62)
- +3.5s: Scv [6] - Smartcmd - Mineralfield750 [50]
- +3.5s: Scv [351] - Move - (19, 61)
- +3.5s: Scv [351] - Smartcmd - Mineralfield [51]
- +3.6s: Scv [160] - Move - (66, 62)
- +3.6s: Scv [160] - Smartcmd - Orbitalcommand [1]
- +3.6s: Scv [154] - Move - (22, 63)
- +3.6s: Scv [154] - Smartcmd - Commandcenter [226]
- +3.6s: Marauder [332] - Attack - (20, 22)
- +3.7s: Barracks [65] - Barrackstrain_marine
- +3.7s: Marauder [347] - Move - (34, 35)
- +3.8s: Mule [305] - Move - (41, 69)
- +3.8s: Mule [305] - Smartcmd - Mineralfield [30]
- +3.8s: Marauder [332] - Move - (40, 36)
- +3.8s: Marauder [347] - Move - (34, 35)
- +3.9s: Marauder [332] - Attack - (20, 22)
- +4.0s: Scv [240] - Move - (65, 63)
- +4.0s: Scv [240] - Smartcmd - Orbitalcommand [1]
- +4.0s: Marauder [347] - Move - (35, 35)
- +4.2s: Marauder [347] - Move - (35, 35)
- +4.2s: Marine [135] - Attack - (36, 57)
- +4.2s: Marine [143] - Attack - (36, 57)
- +4.2s: Marine [147] - Attack - (36, 57)
- +4.2s: Marine [368] - Attack - (17, 41)
- +4.3s: Scv [254] - Move - (24, 66)
- +4.3s: Scv [254] - Smartcmd - Mineralfield [39]
- +4.4s: Scv [311] - Move - (66, 66)
- +4.4s: Scv [311] - Smartcmd - Mineralfield750 [21]
- +4.4s: Scv [337] - Move - (22, 63)
- +4.4s: Scv [337] - Smartcmd - Commandcenter [226]
- +4.4s: Marauder [347] - Move - (35, 35)
- +4.6s: Marauder [332] - Attack - (36, 35)
- +4.6s: Marine [358] - Move - (32, 45)
- +4.6s: Marine [324] - Attack - (37, 21)
- +4.7s: Scv [154] - Move - (21, 66)
- +4.7s: Scv [154] - Smartcmd - Mineralfield [44]
- +4.7s: Marauder [347] - Move - (36, 36)
- +4.7s: Marauder [332] - Attack - (36, 35)
- +4.7s: Marine [358] - Move - (32, 45)
- +4.7s: Marine [324] - Move - (36, 24)
- +4.8s: Scv [160] - Move - (69, 63)
- +4.8s: Scv [160] - Smartcmd - Mineralfield750 [15]
- +4.9s: Scv [10] - Move - (65, 67)
- +4.9s: Scv [10] - Smartcmd - Mineralfield [17]
- +4.9s: Marauder [347] - Move - (36, 36)
- +4.9s: Marauder [332] - Attack - (36, 35)
- +4.9s: Marine [358] - Move - (32, 44)
- +4.9s: Marine [324] - Move - (36, 24)

### Rules for Reasoning && Game Mechanics
Perform the state transition based on the following logic:
1. **Time:** Increment the "Time" field by 5 seconds.
2. **Resources:**
 - Minerals/Gas Change (*$\approx$*) (Income_Rate_per_min / 60 * 5s) - (Cost of new Actions).
 - Note: Income Rate is the (+value) found in the [Info] section.
 - Update Income Rate roughly proportional to the change in active mining workers.
3. **Supply:**
 - Used += Supply Cost of the newly issued Action. (Units in Queue consume Supply, not upon completion.)
 - Cap += Newly Completed Supply Structure (eg. Supply Depot, Command Center).
4. **Queue & Production:**
 - Progress += (5s / Total Build Time at game_speed=1.4). Note: Game Speed is 1.4.
 - If Progress >= 100%: Remove from [Queue], add to [My Units](if unit) or [My Structures](if structure), and update [Info] (Supply Cap, Upgrades) accordingly.
 - New Actions: Add newly issued training/building Action to [Queue].
5. **Movement:** For moving units: New Pos (*$\approx$*) Start Pos + Direction_Vector * (min(Remaining Distance, Speed * 5s)).
6. **Status:** Status Change depends on the Command and the Target.
7. **Combat:**
 - If units are "Attacking" or under attack, reduce HP based on estimated DPS and duration.
 - Remove units if HP <= 0.
8. **Visibility:**
 - Unit Hiding: Units inside structures (e.g., SCVs in Refineries, units in Bunkers/Medivacs) do NOT appear in [My Units] list temporarily, though they still consume Supply.
 - Fog of War: Update [Visible Hostiles] based on current visual radius of self units and structures. Remove enemy units that move into the fog or become cloaked without detection.
9. **Consistency:** Units/Structures not mentioned in the Action Sequence remain unchanged (ID, Position, HP, Status and so on).

### Output Format Requirements
1. **Raw Text Only**: Output the observation text directly. Do NOT use JSON, Markdown code blocks, or conversational fillers (e.g., "Here is the prediction").
2. **Format**: The output must start strictly with the [Info] tag and follow the exact structure of the input observation.

/no_think"
},
\end{lstlisting}
\tcblower
\textbf{Output:}
\begin{lstlisting}[style=promptstyle]
{
    "role": "assistant",
    "content": "<think>

</think>[Info]
Time: 05:49 | Race: Terran | Enemy Race: Terran | Map: Flat64
Minerals: 230 (+2547/min) | Gas: 23 (+268/min)
Supply: 68/93 (Army: 19, Workers: 49)
Alerts: None
Upgrades: Punishergrenades

[Queue]
 - Orbitalcommand [1] at (63,61): Commandcentertrain_scv (75%)
 - Barracks [65] at (53,59): Barrackstrain_marine (7%)
 - Orbitalcommand [83] at (43,64): Commandcentertrain_scv (34%)
 - Barracks [86] at (55,55): Barrackstrain_marine (53%)
 - Barracks [90] at (58,57): Barrackstrain_marine (81%), Barrackstrain_marine (62%)
 - Factory [190] at (43,57): Factorytrain_siegetank (14%)
 - Commandcenter [226] at (24,61): Upgradetoorbital_orbitalcommand (0%)
 - Constructing: Barracksreactor [334] at (65,54) (73%)
 - Constructing: Missileturret [352] at (18,68) (93%)

[My Units]
> Workers: (40Mining:[3,5,6,7,8,9,11,13,35,61,71,77,82,84,115,
139,144,153,155,159,179,183,196,217,222,225,
237,240,254,256,264,271,292,307,311,320,337,
351,360,367], 2Mule:[305,313])
 - Scv [2] at (51,71) (HP:100%, St:Moving)
 - Scv [10] at (65,66) (HP:100%, St:Moving)
 - Scv [154] at (21,66) (HP:100%, St:Moving)
 - Scv [160] at (68,63) (HP:100%, St:Moving)
 - Scv [326] at (17,67) (HP:100%, St:Building)
> Army:
 - Marine [135] at (37,57) (HP:100%, St:Idle)
 - Marine [143] at (37,56) (HP:100%, St:Idle)
 - Marine [147] at (36,57) (HP:100%, St:Idle)
 - Marine [324] at (38,23) (HP:100%, St:Moving)
 - Marauder [332] at (38,35) (HP:73%, St:Attacking)
 - Marauder [347] at (36,36) (HP:21%, St:Moving)
 - Marine [356] at (67,44) (HP:100%, St:Attacking)
 - Marine [358] at (32,46) (HP:100%, St:Moving)
 - Marine [359] at (26,44) (HP:100%, St:Attacking)
 - Marine [368] at (54,61) (HP:100%, St:Attacking)

[My Structures]
 - Orbitalcommand [1] at (63,61) (HP:100%, Eng:16%, St:Training)
 - Supplydepotlowered [36] at (50,60) (HP:100%, St:Idle)
 - Barracks [65] at (53,59) (HP:100%, St:Training, AddOn:Techlab [148])
 - Orbitalcommand [83] at (43,64) (HP:100%, Eng:19%, St:Training)
 - Barracks [86] at (55,55) (HP:100%, St:Training, AddOn:Techlab [150])
 - Barracks [90] at (58,57) (HP:100%, St:Training, AddOn:Reactor [158])
 - Refinery [114] at (70,58) (HP:100%, St:Idle)
 - Refinery [122] at (59,68) (HP:100%, St:Idle)
 - Bunker [140] at (44,54) (HP:100%, St:Idle)
 - Engineeringbay [146] at (57,52) (HP:100%, St:Idle)
 - Barrackstechlab [148] at (56,59) (HP:100%, St:Idle)
 - Barrackstechlab [150] at (58,55) (HP:100%, St:Idle)
 - Supplydepotlowered [156] at (53,57) (HP:100%, St:Idle)
 - Barracksreactor [158] at (61,57) (HP:100%, St:Idle)
 - Supplydepotlowered [164] at (60,53) (HP:100%, St:Idle)
 - Factory [190] at (43,57) (HP:100%, St:Training, AddOn:Techlab [252])
 - Commandcenter [226] at (24,61) (HP:100%, St:Upgrading)
 - Supplydepotlowered [251] at (55,53) (HP:100%, St:Idle)
 - Factorytechlab [252] at (46,57) (HP:100%, St:Idle)
 - Barracks [257] at (62,54) (HP:100%, St:Building)
 - Supplydepotlowered [265] at (42,55) (HP:100%, St:Idle)
 - Missileturret [299] at (43,73) (HP:100%, St:Idle)
 - Missileturret [314] at (70,68) (HP:100%, St:Idle)
 - Supplydepotlowered [333] at (46,59) (HP:100%, St:Idle)
 - Refinery [335] at (28,68) (HP:100%, St:Idle)

[Visible Hostiles]
> Enemy Units:
 - Scv [181] at (34,21) (HP:88%)
 - Banshee [230] at (32,29) (HP:83%, Eng:5%)
 - Banshee [239] at (34,27) (HP:87%, Eng:1%)
 - Vikingassault [281] at (28,36) (HP:57%)
 - Marine [366] at (30,34) (HP:11%)
> Enemy Structures:
 - Barracks [141] at (30,22) (HP:100%)
 - Starport [142] at (30,25) (HP:100%)
 - Starporttechlab [162] at (33,25) (HP:91%)
 - Barracks [306] at (30,28) (HP:100%)
 - Barrackstechlab [365] at (33,22) (HP:92%)
> Snapshot Enemy Structures:
 - Supplydepotlowered at (21,28)
 - Orbitalcommand at (24,22)
 - Supplydepotlowered at (28,28)
 - Factory at (23,28)
 - Refinery at (17,25)
 - Refinery at (28,15)

"
}
\end{lstlisting}
\end{tcolorbox}

\subsection{Example of StarWM-Agent Online Decision Making}\label{Example: StarWM-Agent}
We present an example of the online decision-making context of StarWM-Agent, including the policy's initial action proposal, the world model's simulated future observation, and the refined action produced by the policy. This example demonstrates how the \emph{Generate--Simulate--Refine} loop operates in practice.

\begin{tcolorbox}[
    colback=black!5!white,
    colframe=blue!30!black,
    title=\textbf{Example of StarWM-Agent Online Decision Making},
    fonttitle=\bfseries,
    arc=1mm,
    boxrule=0.8pt,
    width=\linewidth,
    left=2mm,
    right=2mm,
    top=0mm,
    bottom=0mm,
    breakable
]
\begin{lstlisting}[style=promptstyle]
[{
    "role": "user",
    "content": "As a top-tier StarCraft II strategist, your task is to give one or more commands based on the current game state. Only give commands which can be executed immediately, instead of waiting for certain events.

### Aim
Our final aim: destroy all enemies as soon as possible.
Our strategy:
- Resource collection: produce workers and gather minerals and gas
- Development: build attacking units and structures
- Attacking: concentrate forces to search and destroy enemies proactively

### Current Game State
# Round state
Time: 01:02
Race: Terran
Minerals: 145
Vespene: 12
Supply army: 0
Supply workers: 13
Supply unused: 18
Map size: 64x64

# Own units
[577, 145, 281, 713, 857, 457, 1, 721, 569, 425, 865]SCV
State: collecting resources automatically
[289]SCV
Position: (25, 53)
Health: 45/45 (100%)
State: constructing [(24, 53)]

# Unit abilities
SCV[577, 145, 281, 713, 857, 457, 1, 721, 569, 425, 289, 865]: TERRANBUILD_COMMANDCENTER, TERRANBUILD_SUPPLYDEPOT, TERRANBUILD_REFINERY, TERRANBUILD_BARRACKS

# Own structures
[137]CommandCenter
Position: (24, 61)
Health: 1500/1500 (100%)
State: idle
Harvesters: 9/16
[169]SupplyDepot
Position: (24, 57)
Health: 400/400 (100%)
State: idle
[321]Barracks(building 29%)
Position: (24, 53)
[601]Refinery
Position: (17, 58)
Health: 500/500 (100%)
State: idle
Harvesters: 3/3 (no more harvesters accepted)
[745]SupplyDepot
Position: (29, 61)
Health: 400/400 (100%)
State: idle

# Structure abilities
CommandCenter[137]: COMMANDCENTERTRAIN_SCV

# Visible enemy units
[Empty]

# Visible enemy structures
[Empty]

# Action history
{"action": "COMMANDCENTERTRAIN_SCV", "units": [137]}
{"action": "TERRANBUILD_SUPPLYDEPOT", "units": [281], "target_position": [24, 59]}
{"action": "TERRANBUILD_SUPPLYDEPOT", "units": [865], "target_position": [27, 61]}
{"action": "TERRANBUILD_REFINERY", "units": [577], "target_unit": 513}
{"action": "TERRANBUILD_BARRACKS", "units": [289], "target_position": [24, 55]}

# Map information
Closest mineral fields: [665](25, 67), [633](18, 61), [729](24, 68), [801](20, 67), [497](17, 62), [297](18, 65), [193](21, 68), [313](17, 64)
Closest vespene geysers: [513](17, 58), [329](28, 68), [369](36, 65), [945](20, 48)

# Ability description
COMMANDCENTERTRAIN_SCV(target: None): Train a SCV, a Terran worker. Cost: 50 minerals.
TERRANBUILD_BARRACKS(target: Point): Build a Barracks to produce infantry units (Marine, Reaper, Marauder, Ghost). Cost: 150 minerals.
TERRANBUILD_COMMANDCENTER(target: Point): Build a Command Center, the Terran base. Cost: 400 minerals.
TERRANBUILD_REFINERY(target: Unit): Build a Refinery on a Vespene Geyser. Cost: 75 minerals.
TERRANBUILD_SUPPLYDEPOT(target: Point): Build a Supply Depot to increase 8 supplies and unlock Barracks. Cost: 100 minerals.

### Rules
Rule checklist:
1. Produce as many units with the strongest attack power as possible.
2. The total cost of all commands should not exceed the current resources (minerals and gas).
3. Commands should not build redundant structures(e.g. 2 Refinery while one is not fully utilized).
4. Commands should not use abilities that are not supported currently.
5. Commands should not build a structure that is not needed now (e.g. build a Missile Turret but there is no enemy air unit).
6. The unit production list capacity of structures is 5. If the list is full, do not add more units to it.
7. Commands should not send SCV or MULE to gather resources because the system will do it automatically.
8. Commands should not train too many SCVs or MULEs, whose number should not exceed the capacity of CommandCenter and Refinery.
9. Commands can construct a new one Supply Depot only when the remaining unused supply is less than 7.
10. Do not give any action that is irrelevant to the task.
11. Each of units can only be used in the whole response once at most.
12. If a unit is already performing an action as given task, you should ignore it, instead of giving a repeated action for it.
13. If one task cannot be finished, just ignore it.
14. If resource is not enough, just complete the most important part of the task.

Give an action JSON in the following format wrapped with triple backticks:
```
[
    {
        "action": "<action_name>",
        "units": [<unit_id>, <unit_id>, ...], # units you want to command
        "target_unit" (optional): <unit_id>, # some existing unit
        "target_position" (optional): [x, y]
    },
    // more actions ...
]
```

Example:
```
[
    {
        "action": "ATTACK_ATTACK",
        "units": [1, 2, 3],
        "target_unit": 9
    },
    {
        "action": "MOVE_MOVE",
        "units": [4, 5],
        "target_position": [50, 60]
    },
    {
        "action": "COMMANDCENTERTRAIN_SCV",
        "units": [6]
    }
]
```"
},
\end{lstlisting}
\tcbline
\begin{lstlisting}[style=promptstyle]
{
    "role": "assistant",
    "content": "<think>

</think>

```
[
    {
        "action": "TERRANBUILD_SUPPLYDEPOT",
        "units": [713],
        "target_position": [30, 61]
    }
]
```"
},
\end{lstlisting}
\tcbline
\begin{lstlisting}[style=promptstyle]
{
    "role": "user",
    "content": "
Here is a simulation report showing the **PREDICTED GAME STATE** 5 seconds **IF** your original proposed action is executed.

<simulation_report>
# Round state
Time: 01:07
Race: Terran
Enemy Race: Terran
Minerals: 50
Vespene: 24
Supply army: 0
Supply workers: 13
Supply unused: 18
Map size: 64x64

# Own units
[865, 857, 425, 713, 569, 1, 457]Scv
State: collecting resources automatically
[721]Scv
Position: (21, 67)
Health: 100%
State: idle
[281]Scv
Position: (20, 61)
Health: 100%
State: idle
[145]Scv
Position: (24, 65)
Health: 100%
State: idle
[289]Scv
Position: (25, 54)
Health: 100%
State: constructing
[577]Scv
Position: (21, 60)
Health: 100%
State: idle

# Own structures
[137]Commandcenter
Position: (24, 61)
Health: 100%
State: idle
[169]Supplydepot
Position: (24, 57)
Health: 100%
State: idle
[601]Refinery
Position: (17, 58)
Health: 100%
State: idle
[745]Supplydepot
Position: (29, 61)
Health: 100%
State: idle
[321]Barracks
Position: (24, 53)
State: under construction (40%)
[NEW]Supplydepot
Position: (30, 61)
State: under construction (23%)

# Visible enemy units
[Empty]

# Visible enemy structures
[Empty]
</simulation_report>

### TACTICAL ANALYSIS REQUIRED
**CRITICAL INSTRUCTION**: 
1. **LOOK BACK**: You MUST compare the **<simulation_report>** above with the **### Current Game State** provided in the **first user message**.
2. **HYPOTHETICAL STATE**: This report is a **prediction**. The action you proposed previously has **NOT** been executed yet. 
   - **KEEP** your original commands by default.
   - **ADD** new commands if you find idle units/structures or extra resources in the prediction.
   - **REMOVE/CHANGE** original commands ONLY if they lead to bad trades.

Analyze the predicted state and output the OPTIMIZED JSON:

1. **Check Resources & Supply**: 
   - **Supply Check**: Look at "Supply unused" in the prediction. Is it low (< 3)? 
     - If YES, you are about to be Supply Blocked. **ADD** a command to `Build SupplyDepot` (Use a valid SCV).
   - **Gas Strategy**: 
     - Look at "Vespene". Is it very low (e.g., 0 or < 50) while "Minerals" are high? **ADD** `Build Refinery` to start gathering gas.
     - Is "Vespene" high (e.g., > 150)? It's time to develop technology. **ADD** commands to `Build Factory/Starport`, `Build TechLab`, `Upgrade (EngineeringBay)`, or `Train Marauder/SiegeTank`.
   - **Mineral Surplus & Expansion**: 
     - Look at "Minerals". Are they high (e.g., > 400)? It's time to expand. **ADD** Building Orders (e.g., `Build Command Center` (Expand), `Build Barracks` (Increase Production)).

2. **Check Structure States**: 
   - **Queue Limit Check**: Look at the "Production list" of each structure. 
     - Does any list contain 4 or 5 items? (e.g., `Marine (25%), Marine (Waiting), Marine (Waiting)...`).
     - If YES, the queue is **FULL**. **Do NOT** add more training commands to this structure. Instead, if resources allow, **ADD** commands to `Build Barracks/Factory` to increase production capacity.
   - **Fill the Queue**: Look at "Production list". Is it empty or does it only have 1 item?
     - If YES, the structure needs more orders. **ADD** commands to fill the queue (e.g., `Train SCV` for CC, `Train Marine` for Barracks).
   - **Idle Structure Check**: 
     - Look at "Own structures". Are there any marked as **"State: idle"**? If YES, immediately **ADD** commands to train units (e.g., `Train SCV` for CC, `Train Marine` for Barracks).
     - Are tech structures (e.g., TechLab, EngineeringBay) marked as **"State: idle"**? If YES, **ADD** commands to upgrade technology.

3. **Check Unit Health & Survival**: 
   - **Idle Army Check**: Look at "Own units". Are there any Army units marked as **"State: idle"** while the **Visible enemy units** or **Visible enemy structures** list is NOT empty? 
     - If YES, do not just wait. Add an **Attack** command to pressure the enemy.
   - **Survival & Bad Trade Check**: Compare the **Own units** and **Own structures** list in the prediction with the **### Current Game State**. Are any units missing (This means they **DIED**) without Killing or Heavily Damaging enemy units?
     - If YES, this is a **Bad Trade**. **CHANGE** your command to **Retreat** or **Reposition** to save your units.
   - **Damage Check**: Look at the "Health" of your units. Is any unit's health low or dropped significantly?
     - If YES, the unit is under attack. Add a command to **Kite** (move back slightly) or **Heal/Repair** (Medivac/SCV) to save it.
   - **Vision & Scouting Check**: is the **Visible enemy structures** list [Empty]?
     - If YES, your vision is limited. **ADD** `Attack` or `Move` commands to **search for enemy bases**.

### NEW ACTION RULES
   - **VALIDATE ACTORS**: When adding new commands, refer to **Unit abilities**, **Structure abilities**, **Ability description** in the **### Current Game State** provided in the **first user message**. Ensure the selected units actually have the ability to perform the action.
   - **CORRECT UNIT SELECTION**: When adding `TERRANBUILD` commands, you MUST find a valid **SCV** from the **### Current Game State** (marked as `State: idle` or `State: collecting resources automatically`).

**Output Rules**: Create an OPTIMIZED JSON list that combines your beneficial original actions with any NEW necessary and valid actions derived from the analysis above.
/no_think"
},
\end{lstlisting}
\tcbline
\begin{lstlisting}[style=promptstyle]
{
    "role": "assistant",
    "content": "<think>

</think>

```
[
    {
        "action": "COMMANDCENTERTRAIN_SCV",
        "units": [137]
    }
]
```"
}
]
\end{lstlisting}
\end{tcolorbox}

\section{Prompt Templates for World Model and Decision System}

\subsection{World Model Prompt}\label{Prompt:World Model}
We provide the prompt template used for StarWM and zero-shot world models. This prompt is used both in offline training and inference, as well as online simulation within the \emph{Generate--Simulate--Refine} decision loop.

\begin{tcolorbox}[
    colback=black!5!white,
    colframe=blue!30!black,
    title=\textbf{World Model Prompt},
    fonttitle=\bfseries,
    arc=1mm,
    boxrule=0.8pt,
    width=\linewidth,
    left=2mm,
    right=2mm,
    top=0mm,
    bottom=0mm,
    breakable
]
\begin{lstlisting}[style=promptstyle]
### Role
You are the StarCraft II World Model, a dynamics simulator for Player {player_id}.

### Task
Predict the Future Observation for Player {player_id} exactly {delta} seconds later based on the Current Observation, the Action Sequence and the Game Mechanics.

### Input
1. Current Observation: The partial state visible to Player {player_id} at time t.
2. Action Sequence: All actions executed by Player {player_id} between t and t+{delta}s.

### Output
Future Observation: The Predicted partial state visible to Player {player_id} at time t+{delta}s.

### Data Explanation
#### Observation Schema:
- **[Info]**: Player-specific status, race, enemy race, economy (minerals, gas, gather rate), supply (used, cap), alerts and upgrades.
- **[Queue]**: Ongoing and queued production, construction, and research tasks.
- **[My Units]**: Self units (workers & army).
- **[My Structures]**: Self buildings, including key structures for production and technology.
- **[Visible Hostiles]**: Enemy units and structures currently visible within the Fog of War.
#### Action Schema:
The Action follows the format: Subject - Command - Target.
- **Subject**: UnitName [ID] (e.g., Marine [101]).
- **Command**: The specific operation (e.g., Move, Attack, etc.).
- **Target**: Coordinate (x, y) or TargetName [ID] or absent for some commands.

### Input Data
#### Current Observation:
{start_obs}
#### Action Sequence({delta}s):
{action_section}

### Rules for Reasoning && Game Mechanics
Perform the state transition based on the following logic:
1. **Time:** Increment the "Time" field by {delta} seconds.
2. **Resources:**
 - Minerals/Gas Change (*$\approx$*) (Income_Rate_per_min / 60 * {delta}s) - (Cost of new Actions).
 - Note: Income Rate is the (+value) found in the [Info] section.
 - Update Income Rate roughly proportional to the change in active mining workers.
3. **Supply:**
 - Used += Supply Cost of the newly issued Action. (Units in Queue consume Supply, not upon completion.)
 - Cap += Newly Completed Supply Structure (eg. Supply Depot, Command Center).
4. **Queue & Production:**
 - Progress += ({delta}s / Total Build Time at game_speed=1.4). Note: Game Speed is 1.4.
 - If Progress >= 100%: Remove from [Queue], add to [My Units](if unit) or [My Structures](if structure), and update [Info] (Supply Cap, Upgrades) accordingly.
 - New Actions: Add newly issued training/building Action to [Queue].
5. **Movement:** For moving units: New Pos (*$\approx$*) Start Pos + Direction_Vector * (min(Remaining Distance, Speed * {delta}s)).
6. **Status:** Status Change depends on the Command and the Target.
7. **Combat:**
 - If units are "Attacking" or under attack, reduce HP based on estimated DPS and duration.
 - Remove units if HP <= 0.
8. **Visibility:**
 - Unit Hiding: Units inside structures (e.g., SCVs in Refineries, units in Bunkers/Medivacs) do NOT appear in [My Units] list temporarily, though they still consume Supply.
 - Fog of War: Update [Visible Hostiles] based on current visual radius of self units and structures. Remove enemy units that move into the fog or become cloaked without detection.
9. **Consistency:** Units/Structures not mentioned in the Action Sequence remain unchanged (ID, Position, HP, Status and so on).

### Output Format Requirements
1. **Raw Text Only**: Output the observation text directly. Do NOT use JSON, Markdown code blocks, or conversational fillers (e.g., "Here is the prediction").
2. **Format**: The output must start strictly with the [Info] tag and follow the exact structure of the input observation.

\end{lstlisting}
\end{tcolorbox}

\subsection{Online Refinement Prompt with World Model Predictions}\label{Prompt:Online Refinement Prompt with World Model Predictions}
We provide the online refinement prompt used in the \emph{Generate--Simulate--Refine} decision loop of StarWM-Agent. The same refinement prompt is also used in the online zero-shot world model ablation, where StarWM is replaced by a zero-shot Qwen3-8B for future prediction.

\begin{tcolorbox}[
    colback=black!5!white,
    colframe=blue!30!black,
    title=\textbf{Online Refinement Prompt with World Model Predictions},
    fonttitle=\bfseries,
    arc=1mm,
    boxrule=0.8pt,
    width=\linewidth,
    left=2mm,
    right=2mm,
    top=0mm,
    bottom=0mm,
    breakable
]
\begin{lstlisting}[style=promptstyle]
Here is a simulation report showing the **PREDICTED GAME STATE** 5 seconds **IF** your original proposed action is executed.

<simulation_report>
{report}
</simulation_report>

### TACTICAL ANALYSIS REQUIRED
**CRITICAL INSTRUCTION**: 
1. **LOOK BACK**: You MUST compare the **<simulation_report>** above with the **### Current Game State** provided in the **first user message**.
2. **HYPOTHETICAL STATE**: This report is a **prediction**. The action you proposed previously has **NOT** been executed yet. 
   - **KEEP** your original commands by default.
   - **ADD** new commands if you find idle units/structures or extra resources in the prediction.
   - **REMOVE/CHANGE** original commands ONLY if they lead to bad trades.

Analyze the predicted state and output the OPTIMIZED JSON:

1. **Check Resources & Supply**: 
   - **Supply Check**: Look at "Supply unused" in the prediction. Is it low (< 3)? 
     - If YES, you are about to be Supply Blocked. **ADD** a command to `Build SupplyDepot` (Use a valid SCV).
   - **Gas Strategy**: 
     - Look at "Vespene". Is it very low (e.g., 0 or < 50) while "Minerals" are high? **ADD** `Build Refinery` to start gathering gas.
     - Is "Vespene" high (e.g., > 150)? It's time to develop technology. **ADD** commands to `Build Factory/Starport`, `Build TechLab`, `Upgrade (EngineeringBay)`, or `Train Marauder/SiegeTank`.
   - **Mineral Surplus & Expansion**: 
     - Look at "Minerals". Are they high (e.g., > 400)? It's time to expand. **ADD** Building Orders (e.g., `Build Command Center` (Expand), `Build Barracks` (Increase Production)).

2. **Check Structure States**: 
   - **Queue Limit Check**: Look at the "Production list" of each structure. 
     - Does any list contain 4 or 5 items? (e.g., `Marine (25%), Marine (Waiting), Marine (Waiting)...`).
     - If YES, the queue is **FULL**. **Do NOT** add more training commands to this structure. Instead, if resources allow, **ADD** commands to `Build Barracks/Factory` to increase production capacity.
   - **Fill the Queue**: Look at "Production list". Is it empty or does it only have 1 item?
     - If YES, the structure needs more orders. **ADD** commands to fill the queue (e.g., `Train SCV` for CC, `Train Marine` for Barracks).
   - **Idle Structure Check**: 
     - Look at "Own structures". Are there any marked as **"State: idle"**? If YES, immediately **ADD** commands to train units (e.g., `Train SCV` for CC, `Train Marine` for Barracks).
     - Are tech structures (e.g., TechLab, EngineeringBay) marked as **"State: idle"**? If YES, **ADD** commands to upgrade technology.

3. **Check Unit Health & Survival**: 
   - **Idle Army Check**: Look at "Own units". Are there any Army units marked as **"State: idle"** while the **Visible enemy units** or **Visible enemy structures** list is NOT empty? 
     - If YES, do not just wait. Add an **Attack** command to pressure the enemy.
   - **Survival & Bad Trade Check**: Compare the **Own units** and **Own structures** list in the prediction with the **### Current Game State**. Are any units missing (This means they **DIED**) without Killing or Heavily Damaging enemy units?
     - If YES, this is a **Bad Trade**. **CHANGE** your command to **Retreat** or **Reposition** to save your units.
   - **Damage Check**: Look at the "Health" of your units. Is any unit's health low or dropped significantly?
     - If YES, the unit is under attack. Add a command to **Kite** (move back slightly) or **Heal/Repair** (Medivac/SCV) to save it.
   - **Vision & Scouting Check**: is the **Visible enemy structures** list [Empty]?
     - If YES, your vision is limited. **ADD** `Attack` or `Move` commands to **search for enemy bases**.

### NEW ACTION RULES
   - **VALIDATE ACTORS**: When adding new commands, refer to **Unit abilities**, **Structure abilities**, **Ability description** in the **### Current Game State** provided in the **first user message**. Ensure the selected units actually have the ability to perform the action.
   - **CORRECT UNIT SELECTION**: When adding `TERRANBUILD` commands, you MUST find a valid **SCV** from the **### Current Game State** (marked as `State: idle` or `State: collecting resources automatically`).

**Output Rules**: Create an OPTIMIZED JSON list that combines your beneficial original actions with any NEW necessary and valid actions derived from the analysis above.
\end{lstlisting}
\end{tcolorbox}

\subsection{Online Refinement Prompt for Self-Reflection Ablation}\label{Prompt:Online Refinement Prompt for Self-Reflection Ablation}
We provide the refinement prompt used in the online self-reflection ablation study. In this setting, no external future prediction is introduced, and the policy refines its initially generated action solely based on its own internal reasoning.

\begin{tcolorbox}[
    colback=black!5!white,
    colframe=blue!30!black,
    title=\textbf{Online Refinement Prompt for Self-Reflection Ablation},
    fonttitle=\bfseries,
    arc=1mm,
    boxrule=0.8pt,
    width=\linewidth,
    left=2mm,
    right=2mm,
    top=0mm,
    bottom=0mm,
    breakable
]
\begin{lstlisting}[style=promptstyle]
Here is a reflection guidance for your original proposed actions.
### TACTICAL REFLECTION REQUIRED
**CRITICAL INSTRUCTION**: 
1. **LOOK BACK**: You MUST review the **### Current Game State** provided in the **first user message**.
2. **REFLECTION**: The actions you proposed previously have **NOT** been executed yet. You have a chance to refine it.
   - **KEEP** your original commands by default.
   - **ADD** new commands if you find idle units/structures or extra resources in the current state.
   - **REMOVE/CHANGE** original commands ONLY if they are not good enough.

Analyze the current state and output the OPTIMIZED JSON:

1. **Check Resources & Supply**: 
   - **Supply Check**: Look at "Supply unused". Is it low (< 3)? 
     - If YES, you are about to be Supply Blocked. **ADD** a command to `Build SupplyDepot` (Use a valid SCV).
   - **Gas Strategy**: 
     - Look at "Vespene". Is it very low (e.g., 0 or < 50) while "Minerals" are high? **ADD** `Build Refinery` to start gathering gas.
     - Is "Vespene" high (e.g., > 150)? It's time to develop technology. **ADD** commands to `Build Factory/Starport`, `Build TechLab`, `Upgrade (EngineeringBay)`, or `Train Marauder/SiegeTank`.
   - **Mineral Surplus & Expansion**: 
     - Look at "Minerals". Are they high (e.g., > 400)? It's time to expand. **ADD** Building Orders (e.g., `Build Command Center` (Expand), `Build Barracks` (Increase Production)).

2. **Check Structure States**: 
   - **Queue Limit Check**: Look at the "Production list" of each structure. 
     - Does any list contain 4 or 5 items?
     - If YES, the queue is **FULL**. **Do NOT** add more training commands to this structure. Instead, if resources allow, **ADD** commands to `Build Barracks/Factory` to increase production capacity.
   - **Fill the Queue**: Look at "Production list". Is it empty or does it only have 1 item?
     - If YES, the structure needs more orders. **ADD** commands to fill the queue (e.g., `Train SCV` for CC, `Train Marine` for Barracks).
   - **Idle Structure Check**: 
     - Look at "Own structures". Are there any marked as **"State: idle"**? If YES, immediately **ADD** commands to train units (e.g., `Train SCV` for CC, `Train Marine` for Barracks).
     - Are tech structures (e.g., TechLab, EngineeringBay) marked as **"State: idle"**? If YES, **ADD** commands to upgrade technology.

3. **Check Unit Health & Status**: 
   - **Idle Army Check**: Look at "Own units". Are there any Army units marked as **"State: idle"** while the **Visible enemy units** or **Visible enemy structures** list is NOT empty? 
     - If YES, do not just wait. Add an **Attack** command to pressure the enemy.
   - **Damage Check**: Look at the "Health" of your units. Is any unit's health low?
     - If YES, add a command to **Kite** (move back slightly) or **Heal/Repair** (Medivac/SCV) to save it.
   - **Vision & Scouting Check**: Is the **Visible enemy structures** list [Empty]?
     - If YES, your vision is limited. **ADD** `Attack` or `Move` commands to **search for enemy bases**.

### NEW ACTION RULES
   - **VALIDATE ACTORS**: When adding new commands, refer to **Unit abilities**, **Structure abilities**, **Ability description** in the **### Current Game State** provided in the **first user message**. Ensure the selected units actually have the ability to perform the action.
   - **CORRECT UNIT SELECTION**: When adding `TERRANBUILD` commands, you MUST find a valid **SCV** from the **### Current Game State** (marked as `State: idle` or `State: collecting resources automatically`).

**Output Rules**: Create an OPTIMIZED JSON list that combines your beneficial original actions with any NEW necessary and valid actions derived from the reflection above.
\end{lstlisting}
\end{tcolorbox}

\end{document}